\documentclass{article}

\PassOptionsToPackage{numbers}{natbib}
\usepackage[preprint]{neurips_2026}

\usepackage[utf8]{inputenc} 
\usepackage[T1]{fontenc}    
\usepackage{hyperref}       
\usepackage{url}            
\usepackage{booktabs}       
\usepackage{longtable}
\usepackage{amsmath}
\usepackage{amsfonts}       
\usepackage{nicefrac}       
\usepackage{microtype}      
\usepackage{xcolor}         
\usepackage{colortbl}       

\usepackage{enumitem}
\usepackage{booktabs}
\usepackage{multirow}
\usepackage{graphicx}
\usepackage{wrapfig}       
\usepackage[normalem]{ulem}
\useunder{\uline}{\ul}{}

\usepackage{tcolorbox}

\title{LaTER: Efficient Test-Time Reasoning \\
via Latent Exploration and Explicit Verification}

\author{
  \textbf{Xuan Li\textsuperscript{1}} \quad
  \textbf{Yining Wang\textsuperscript{2}} \quad
  \textbf{Yuchen Liu} \quad
  \textbf{Guanjun Liu\textsuperscript{1}} \quad
  \textbf{Delai Qiu\textsuperscript{2}} \\
  \textbf{Shengping Liu\textsuperscript{2}} \quad
  \textbf{Jiaen Liang\textsuperscript{2}} \quad
  \textbf{Wei Huang\textsuperscript{2}} \quad
  \textbf{Jun Yu\textsuperscript{1}\footnotemark[2]} \quad
  \textbf{Junnan Zhu\textsuperscript{3}\footnotemark[2]}
  \\
  \textsuperscript{1}University of Science and Technology of China, 
  \\
  \textsuperscript{2}Unisound AI Technology Co., Ltd,
  \\
  \textsuperscript{3}MAIS, Institute of Automation, Chinese Academy of Sciences
  \\
  {\fontsize{11pt}{0pt}\selectfont
    \texttt{harryjun@ustc.edu.cn, junnan.zhu@nlpr.ia.ac.cn}
  }
}

\begin{document}

\maketitle
\begingroup
\renewcommand{\thefootnote}{\fnsymbol{footnote}}
\footnotetext[2]{Corresponding author.}
\endgroup

\begin{abstract}
  Chain-of-thought (CoT) reasoning improves large language models (LLMs) on difficult tasks, but it also makes inference expensive because every intermediate step must be generated as a discrete token. Latent reasoning reduces visible token generation by propagating continuous states, yet replacing explicit derivations with latent computation can hurt tasks that require symbolic checking. We propose \textbf{La}tent-\textbf{T}hen-\textbf{E}xplicit \textbf{R}easoning (\textbf{LaTER}), a two-stage paradigm that first performs bounded exploration in a continuous latent space and then switches to explicit CoT for verification and answer generation. In a training-free instantiation, LaTER projects final-layer hidden states back to the input embedding space, preserves the latent KV cache, and uses entropy and model-native stop-token probes to decide when to switch. We find that strong reasoning models already exhibit structured latent trajectories under this interface. On Qwen3-14B, training-free LaTER reduces total token usage by 16\%--32\% on several benchmarks while matching or improving accuracy on most of them; for example, it improves AIME 2025 from 70.0\% to 73.3\% while reducing tokens from 15,730 to 10,661. We further construct \textsc{Latent-Switch-69K}, a supervised corpus that pairs condensed solution intuitions with shortened explicit derivations. Fine-tuning with latent rollout and halting supervision yields additional gains: trained LaTER reaches 80.0\% accuracy on AIME 2025, 10.0 points above the standard CoT baseline, while using 33\% fewer tokens. Our code, data, and model are available at \url{https://github.com/TioeAre/LaTER}.
\end{abstract}

\section{Introduction}

CoT prompting is a simple and effective way to improve reasoning in LLMs~\citep{wei2022chain}. By generating intermediate derivations before the final answer, CoT improves performance on mathematics, science, and code tasks~\citep{guo2025deepseek}. Its main drawback is cost. Strong reasoning models often produce long visible traces, and each additional token increases latency, memory traffic, and attention computation~\citep{vaswani2017attention}. The cost is especially high when the model spends many tokens on tentative exploration, syntactic scaffolding, or discarded solution paths before reaching a stable derivation.

Recent work therefore studies reasoning in a continuous latent space~\citep{hao2025training, zhang2026soft}. Instead of sampling a visible token at every reasoning step, a model can feed back a hidden state or a soft embedding as the next input, using either an analytic mapping such as a pseudo-inverse projection~\citep{zou2025latentcollaborationmultiagentsystems} or a learned projector~\citep{xu2025softcot}, and only decodes discrete readable tokens in the final answer stage. This can substantially reduce visible token generation and has shown promising efficiency gains~\citep{hao2025training, yu2026latentspacefoundationevolution, zhu2025surveylatentreasoning}. However, pure latent reasoning also has a clear weakness: when a problem requires careful symbolic manipulation, explicit checking, or exact answer formatting, fully replacing CoT with latent computation can reduce accuracy on difficult benchmarks such as MATH-500 and AIME~\citep{deng2025latent, tan2026think, shen-etal-2025-codi}.

This suggests that latent and discrete reasoning should not be viewed as mutually exclusive alternatives. A more natural division of labor is to use continuous computation for early exploration and reserve discrete tokens for verification. Human solvers often behave in a similar way: they may first search mentally for a plan and only later write a step-by-step solution. We use this analogy only as motivation for a computational design. The central question is whether an LLM can spend part of its test-time computation in a high-bandwidth latent state, then return to explicit CoT when precise symbolic reasoning is most valuable.

We propose \textbf{LaTER} (\textbf{La}tent-\textbf{T}hen-\textbf{E}xplicit \textbf{R}easoning), a hybrid reasoning paradigm that separates exploration from verification. Given a prompt, LaTER first performs a bounded latent rollout. At each latent step, the final-layer hidden state is mapped back into the input embedding space and reused as the next input, without committing to a visible token. The model then switches to ordinary token generation while preserving the latent-phase KV cache, so the explicit derivation is conditioned on the preceding latent trajectory rather than starting from scratch.

We study LaTER in two settings. First, we show that no additional training is required for the interface to be useful. A training-free version uses a simple adaptive switch based on latent entropy and decoded stop-token probes. On Qwen3-14B, this already improves AIME 2025 from 70.0\% to 73.3\% while reducing average token usage from 15,730 to 10,661, and improves MATH-500 from 93.4\% to 97.2\% with 17\% fewer tokens. Second, we train a LaTER model on \textsc{Latent-Switch-69K}, a dataset designed to teach the model how to allocate latent exploration before explicit reasoning. The trained model reaches 80.0\% on AIME 2025, a 10.0-point gain over the standard CoT baseline, while using 33\% fewer tokens.

Our contributions are threefold. (i) We introduce a latent-then-explicit reasoning interface that preserves the latent KV cache and turns latent computation into a precursor to explicit verification. (ii) We identify training-free latent switching signals, including terminating-token probes and entropy dynamics, showing that pretrained reasoning models can already support structured latent rollouts. (iii) We construct \textsc{Latent-Switch-69K} and train a LaTER model that improves the accuracy--efficiency tradeoff across mathematics, coding, and knowledge-intensive reasoning benchmarks.

\section{Training-Free LaTER}
\label{sec:training_free_later}

We first ask whether a pretrained reasoning model can benefit from a latent-first, explicit-second procedure without any task-specific training. This setting isolates the inference-time interface from supervised adaptation. We show that strong reasoning models can perform several continuous latent steps, retain those steps in the KV cache, and then convert the accumulated state into explicit CoT with lower token usage. We also show that fixed latent horizons are brittle, motivating adaptive switching based on the model's own latent dynamics.

\subsection{Preliminaries and notation.}
\label{subsec:background_later}

\begin{figure}[t]
  \centering
  \vspace{-1.5em}
  \includegraphics[width=\textwidth]{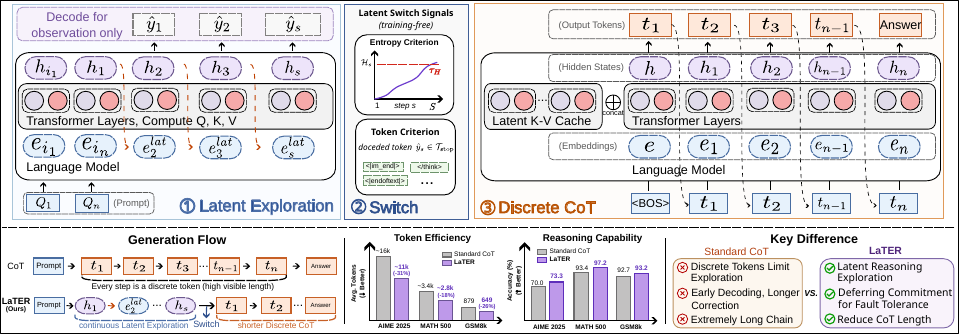}
  \vspace{-1.5em}
  \caption{\textbf{Overview of training-free LaTER.} Given a user prompt, the model first enters a latent reasoning phase, where the final-layer hidden state is projected back into the input embedding space and reused as the next-step input, without committing to visible tokens. The model then switches to explicit CoT decoding, reusing the latent KV cache to generate reasoning steps and the final answer.}
  \label{fig:training_free_later_overview}
  \vspace{-1.5em}
\end{figure}

Let $Q=(Q_1,\ldots,Q_m)$ denote the prompt. At latent step $s$, the model produces a final-layer hidden state $h_s\in\mathbb{R}^{d_h}$. Instead of decoding $h_s$ to a token ID and feeding that token back into the model, we map $h_s$ directly into the input embedding space. Following the latent-transition construction of LatentMAS~\citep{zou2025latentcollaborationmultiagentsystems}, we use
\begin{equation}
  e_{s+1}^{\mathrm{lat}} = W_a h_s,
  \qquad
  W_a \approx W_{out}^{\dagger} W_{in},
  \label{eq:latent_projection}
\end{equation}
where $W_{in}$ is the input embedding matrix, $W_{out}$ is the output projection matrix, and $W_{out}^{\dagger}$ denotes the pseudo-inverse of $W_{out}$. The vector $e_{s+1}^{\mathrm{lat}}$ is then used as the next-step input embedding. This produces a continuous trajectory
\begin{equation}
  h_1 \rightarrow e_2^{\mathrm{lat}} \rightarrow h_2 \rightarrow e_3^{\mathrm{lat}} \rightarrow \cdots \rightarrow h_S,
\end{equation}
with no discrete token commitment at the intermediate latent positions. For diagnostics only, we decode each latent hidden state into a probe distribution and an argmax probe token,
\begin{equation}
  p_s = \mathrm{softmax}(W_{out} h_s),
  \qquad
  \hat{y}_s = \arg\max_i (p_s(i)),
  \label{eq:latent_decode}
\end{equation}
The probe token $\hat{y}_s$ is never used as the next input. It is only an observation of how the latent state aligns with the model's vocabulary space. We also compute the entropy of the probe distribution,
\begin{equation}
  \mathcal{H}_s = - \sum_i p_s(i)\log p_s(i).
  \label{eq:entropy}
\end{equation}
which provides a scalar summary of the model's uncertainty at that latent step.

After the latent rollout, LaTER switches to ordinary explicit CoT decoding. The switch is not a reset: we pass the latent-phase \texttt{past\_key\_values} into the explicit phase, so the generated derivation conditions on the latent trajectory. We evaluate two switching policies:
\begin{itemize}[leftmargin=10pt]
  \item \textbf{Fixed-step switching.} The model performs $N$ latent steps and then enters explicit CoT decoding.
  \item \textbf{Adaptive switching.} The model exits latent reasoning when either the entropy crosses a threshold, the decoded probe token belongs to a model-specific set of terminating tokens such as \texttt{<|im\_end|>}, \texttt{</think>}, or \texttt{<|endoftext|>}.
\end{itemize}

Formally, the adaptive switch is
\begin{equation}
  \mathrm{switch}(s)=
  \mathbf{1}\!\left[
    \mathcal{H}_s > \tau_{\mathcal{H}}
    \;\;\lor\;\;
    \hat{y}_s \in \mathcal{T}_{\mathrm{stop}}
  \right],
  \label{eq:switch_rule}
\end{equation}
where $\tau_{\mathcal{H}}$ is an entropy threshold and $\mathcal{T}_{\mathrm{stop}}$ is the terminating-token set. The next subsection explains why these two signals are empirically meaningful.

\subsection{Empirical motivation: latent trajectories are structured}
\label{subsec:motivation}

A concern with latent reasoning is that hidden states might drift away from the vocabulary manifold, making repeated latent transitions unstable or semantically meaningless. Our experiments suggest a more structured picture for reasoning models such as Qwen3-14B~\citep{yang2025qwen3technicalreport}, DeepSeek-R1-Distill-Llama-8B~\citep{guo2025deepseek}, and OLMo3-32B-Think~\citep{olmo2026olmo3}.

\begin{wrapfigure}{r}{0.45\textwidth}
  \centering
  \vspace{-2.5em}
  \includegraphics[width=0.45\textwidth]{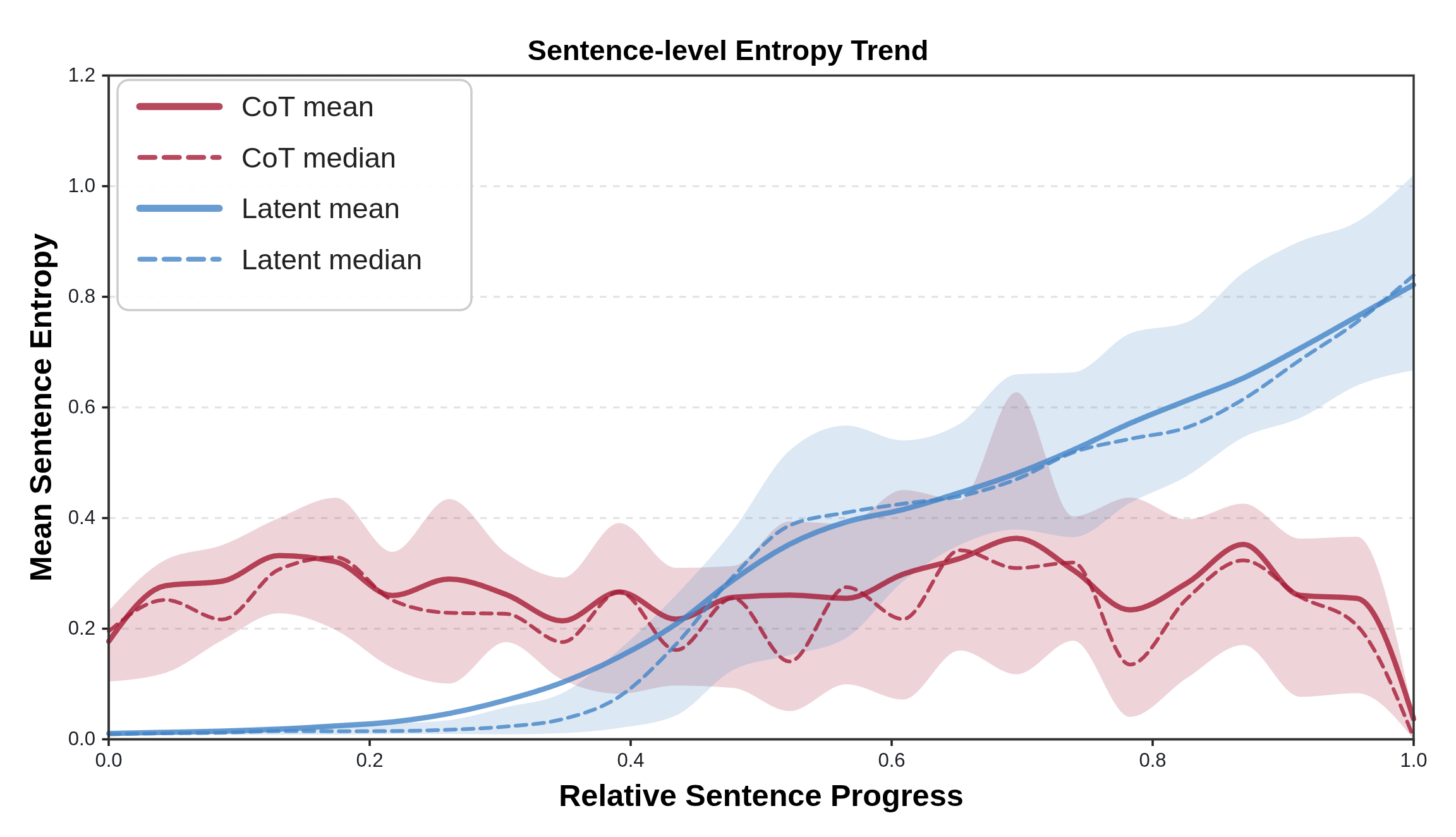}
  \vspace{-2em}
  \caption{Entropy over normalized reasoning progress on AIME 2025 for Qwen3-14B. Blue: mean latent-reasoning entropy after aligning each example from latent start to end. Red: mean CoT entropy after normalizing each sentence by within-sentence progress.}
  \vspace{-1.5em}
  \label{fig:sentence_mean_entropy_trend}
\end{wrapfigure}

\textbf{Phenomenon 1: probe tokens reveal autoregressive stopping structure.}
Early latent states often decode to low-content probes, such as empty strings or repeated newline symbols (\texttt{"\textbackslash n\textbackslash n"}). After additional latent steps, however, the argmax probe frequently reaches model-native terminating symbols such as \texttt{<|im\_end|>}, \texttt{</think>}, or \texttt{<|endoftext|>}. These probe tokens are not fed back into the model, so they do not drive the rollout. Their appearance instead indicates that the continuous trajectory remains coupled to the model's generative prior. In this sense, latent reasoning does not behave like arbitrary numerical drift; it often approaches states that the language model itself would interpret as closure.

This observation is central to LaTER. If the model internally approaches a state that resembles ``ready to stop'', then switching to explicit CoT can be aligned with the model's own trajectory and the reasoning patterns it acquired during pretraining, rather than imposed at an unrelated time.

\textbf{Phenomenon 2: entropy supports an explore-then-verify interpretation.}
As shown in Figure~\ref{fig:sentence_mean_entropy_trend}, the average entropy during latent rollout tends to rise over normalized latent progress before termination. This differs from ordinary explicit decoding, where entropy is often locally high near the beginning of a sentence and then declines as syntax and previously generated words constrain the continuation. The latent phase therefore appears to support a broader and less locally constrained search, while the later explicit phase converts the accumulated state into a step-by-step derivation.

We do not claim that entropy alone fully explains latent reasoning. Rather, these two observations provide practical switching signals: the terminating-token probe suggests that the trajectory is approaching closure, and the entropy profile indicates when the latent state is entering a high-uncertainty regime. Together they motivate the adaptive rule in Eq.~\ref{eq:switch_rule}.

\subsection{Training-free experimental setup}
\label{subsec:exp_setup_training_free}

We compare standard discrete CoT decoding with training-free LaTER under the same prompts and decoding settings. We report accuracy and total token usage. For LaTER, token usage counts both latent steps and emitted explicit tokens, so reductions are not an artifact of ignoring latent computation. We evaluate Qwen3-14B, DeepSeek-R1-Distill-Llama-8B, and OLMo3-32B-Think on AIME 2025~\citep{balunovic_srimatharena_2025}, MATH-500~\citep{lightman2023lets}, GSM8K~\citep{cobbe2021trainingverifierssolvemath}, GPQA~\citep{rein2024gpqa}, ARC-Challenge~\citep{allenai_arc}, HumanEval+, and MBPP+~\citep{evalplus, evalperf}.

\subsection{Fixed-steps switching results}
\label{subsec:fixed_steps_results_training_free}

\begin{wrapfigure}{r}{0.45\textwidth}
  \centering
  \vspace{-2.5em}
  \includegraphics[width=0.45\textwidth]{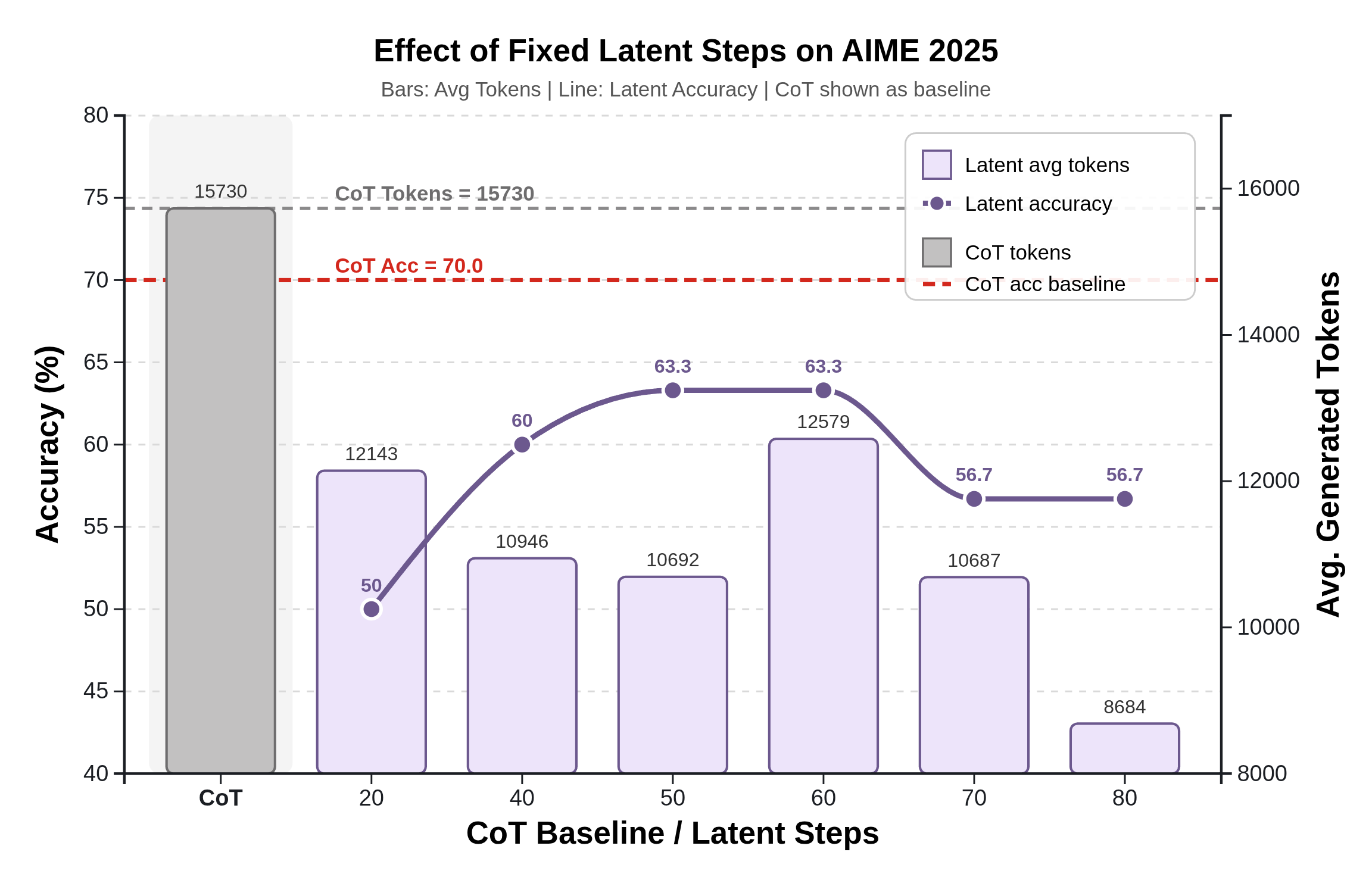}
  \vspace{-2em}
  \caption{Accuracy and token usage on AIME 2025 as the fixed latent-step budget varies.}
  \label{fig:fixed_latent_steps_aime}
  \vspace{-2em}
\end{wrapfigure}

For Qwen3-14B, we follow the official decoding recommendations: temperature $=0.6$, top-$p=0.95$, top-$k=20$, and \texttt{max\_new\_tokens}$=38192$. Under this setup, the standard discrete CoT baseline reaches 70.0\% accuracy on AIME 2025 with roughly $16\mathrm{K}$ tokens on average. Figure~\ref{fig:fixed_latent_steps_aime} shows that fixed-step LaTER can reduce token usage substantially, but it does not fully match the baseline accuracy. The best fixed horizons, around 50--60 latent steps, reach 63.3\% accuracy with about $10\mathrm{K}$--$12\mathrm{K}$ total tokens.

The fixed-step curve is non-monotonic: performance first improves as the latent budget increases, then degrades when the return to explicit reasoning is delayed too long. This pattern supports the role separation behind LaTER. Latent exploration is useful up to a point, but difficult problems still benefit from an explicit symbolic phase that checks intermediate conclusions and formats the final answer. A single fixed horizon cannot adapt to instance difficulty, which motivates the adaptive switch.

\subsection{Adaptive switching results}
\label{subsec:adaptive_results_training_free}

Adaptive LaTER uses the same decoding configuration as above but replaces the fixed latent horizon with Eq.~\ref{eq:switch_rule}. At each latent step, we monitor the entropy of the probe distribution and the argmax probe token. The model switches to explicit CoT once the entropy exceeds $7$ or the probe token becomes a terminating symbol such as \texttt{<|im\_end|>}. Table~\ref{tab:training_free_benchmark} shows that this simple policy usually reduces token usage and often preserves or improves accuracy relative to the paired CoT baseline.
These gains are particularly evident in stronger models and on tasks requiring extended reasoning, achieve greater token savings without compromising solution quality.
The effect is strongest for Qwen3-14B: adaptive LaTER improves AIME 2025 from 70.0\% to 73.3\% while reducing tokens from 15,730 to 10,661, and it improves MATH-500 from 93.4\% to 97.2\% while reducing tokens by 17\%.

\begin{figure}[t]
  \centering
  \vspace{-1em}
  \includegraphics[width=\linewidth]{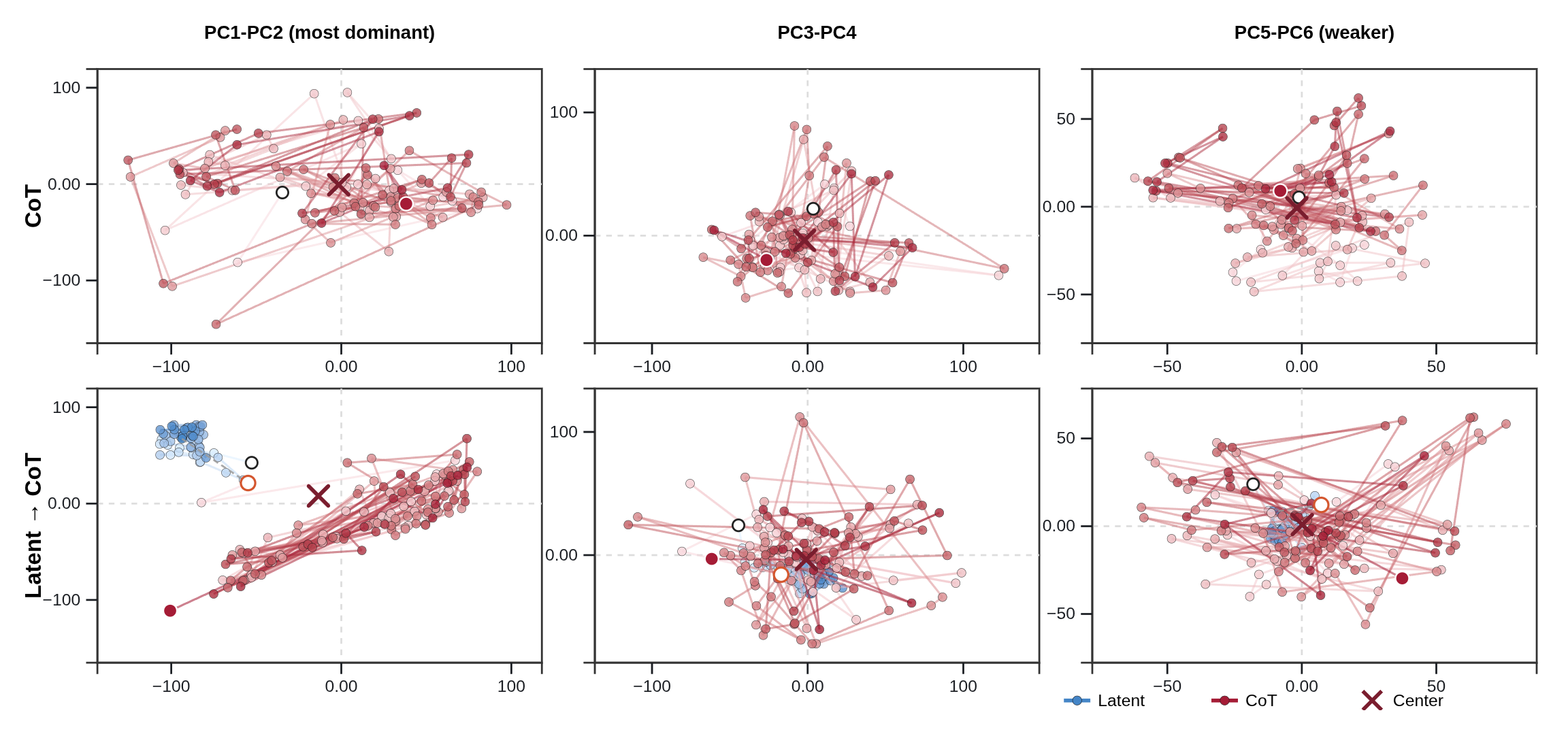}
  \vspace{-2.5em}
  \caption{PCA trajectories of Qwen3-14B on an AIME 2025 example under adaptive switching. The plotted coordinates are the first six PCA components of the final-layer hidden states. Blue denotes the latent trajectory, and red denotes the first 256 steps of the explicit CoT trajectory. Color intensity increases from light to dark as reasoning progresses.}
  \label{fig:pca_trajectory_figure}
  \vspace{-1.5em}
\end{figure}

Figure~\ref{fig:pca_trajectory_figure} gives a qualitative view on one AIME 2025 example. In the dominant PC1--PC2 plane, the discrete CoT trajectory is relatively scattered, suggesting that the model is still searching for a solution path. LaTER first follows a compact latent trajectory. After switching, its explicit trajectory forms repeated refinements along a shared direction rather than spreading randomly. This visualization is not a proof of mechanism, but it is consistent with the hypothesis that latent exploration organizes the state before explicit verification.

\definecolor{accgain}{RGB}{214,239,223}
\definecolor{tokgain}{RGB}{221,233,248}
\definecolor{mildloss}{RGB}{248,228,228}
\newcommand{\accup}[2]{\cellcolor{accgain}\textbf{#1} {\scriptsize(+#2)}}
\newcommand{\tokdown}[2]{\cellcolor{tokgain}\textbf{#1} {\scriptsize($\downarrow$#2\%)}}
\newcommand{\tokup}[2]{\cellcolor{mildloss}#1 {\scriptsize($\uparrow$#2\%)}}
\newcommand{\later}{\textbf{LaTER}}
\begin{table}[t]
  \centering
  \small
  \setlength{\tabcolsep}{3.2pt}
  \renewcommand{\arraystretch}{1.08}
  \caption{Training-free LaTER with adaptive switching across seven benchmarks, compared with the corresponding discrete CoT baseline for each backbone. Token counts include latent steps plus emitted explicit tokens. Green cells mark accuracy gains over the paired baseline, blue cells mark token reductions, and red cells mark token increases.}
  \label{tab:training_free_benchmark}
  \resizebox{\columnwidth}{!}{%
    \begin{tabular}{@{}ll@{\hspace{4pt}}cc@{\hspace{5pt}}cc@{\hspace{5pt}}cc@{\hspace{5pt}}cc@{\hspace{5pt}}cc@{\hspace{5pt}}cc@{\hspace{5pt}}cc@{}}
      \toprule
      \multirow{2}{*}{Model} & \multirow{2}{*}{Method}
      & \multicolumn{2}{c}{AIME 2025}
      & \multicolumn{2}{c}{GSM8K}
      & \multicolumn{2}{c}{MATH-500}
      & \multicolumn{2}{c}{ARC-Challenge}
      & \multicolumn{2}{c}{GPQA}
      & \multicolumn{2}{c}{HumanEval+}
      & \multicolumn{2}{c}{MBPP+} \\
      \cmidrule(lr){3-4}\cmidrule(lr){5-6}\cmidrule(lr){7-8}\cmidrule(lr){9-10}\cmidrule(lr){11-12}\cmidrule(lr){13-14}\cmidrule(lr){15-16}
      & & Acc. & Token & Acc. & Token & Acc. & Token & Acc. & Token & Acc. & Token & Acc. & Token & Acc. & Token \\
      \midrule
      \multirow{2}{*}{DS-Distill-Llama-8B}
      & Baseline & 33.3 & 18546 & 76.7 & 498 & 82.6 & 2932 & 87.7 & 844 & 39.9 & 6186 & 76.2 & 3890 & 67.1 & 3154 \\
      & \later   & 30.0 & \tokdown{11622}{37} & \accup{78.6}{1.9} & \tokdown{378}{24} & 79.8 & 2927 & 85.6 & \tokup{983}{17} & \accup{41.9}{2.0} & \tokdown{5896}{5} & 73.1 & \tokdown{3051}{22} & \accup{69.3}{2.2} & \tokdown{2351}{25} \\
      \midrule
      \multirow{2}{*}{Qwen3-14B}
      & Baseline & 70.0 & 15730 & 92.7 & 879 & 93.4 & 3472 & 96.3 & 631 & 62.6 & 6301 & 90.8 & 2427 & 82.0 & 2144 \\
      & \later   & \accup{73.3}{3.3} & \tokdown{10661}{32} & \accup{93.2}{0.5} & \tokdown{649}{26} & \accup{97.2}{3.8} & \tokdown{2887}{17} & \accup{96.5}{0.2} & \tokdown{533}{16} & 60.6 & \tokdown{4445}{29} & 90.8 & \tokdown{1790}{26} & 79.6 & \tokdown{1760}{18} \\
      \midrule
      \multirow{2}{*}{OLMo3-32B-Think}
      & Baseline & 60.0 & 17042 & 94.3 & 971 & 97.6 & 4197 & 95.0 & 791 & 61.6 & 10460 & 89.0 & 3260 & 82.5 & 2760 \\
      & \later   & 60.0 & \tokdown{13251}{22} & \accup{96.5}{2.2} & 882 & 93.6 & \tokdown{3427}{18} & \accup{95.3}{0.3} & \tokup{882}{12} & 60.1 & \tokdown{7145}{32} & 89.0 & 3201 & 80.9 & 2710 \\
      \bottomrule
    \end{tabular}%
  }
\end{table}

\begin{wrapfigure}{r}{0.45\textwidth}
  \centering
  \vspace{-1.5em}
  \includegraphics[width=\linewidth]{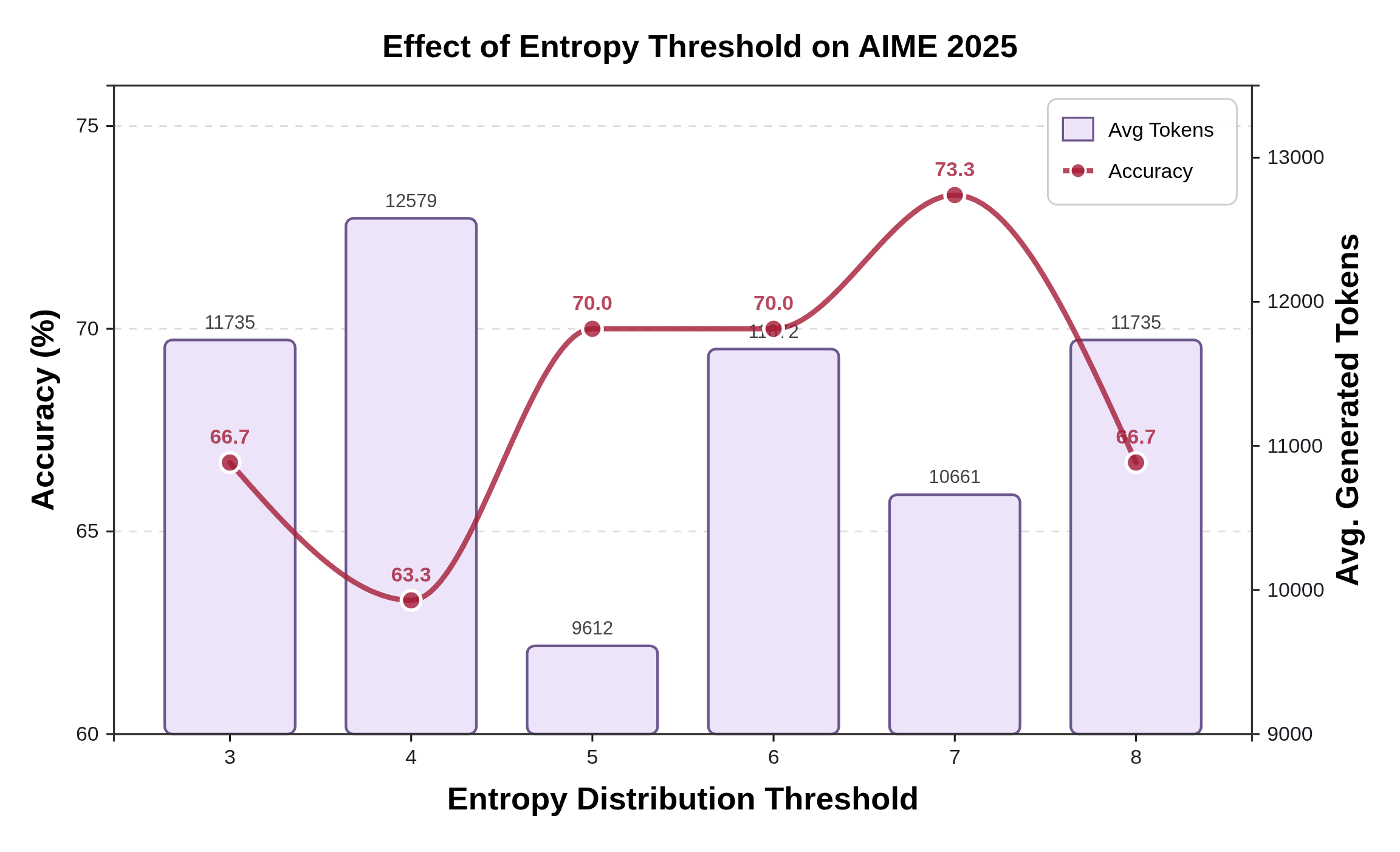}
  \vspace{-2em}
  \caption{Effect of the entropy threshold $\tau_{\mathcal{H}}$ on training-free adaptive LaTER for Qwen3-14B on AIME 2025}
  \label{fig:entropy_threshold_combo_chart}
  \vspace{-1.0em}
\end{wrapfigure}

\textbf{Why does adaptive switching help?}
A fixed horizon gives every instance the same latent budget, regardless of difficulty or internal confidence. Figures~\ref{fig:fixed_latent_steps_aime} and Figures~\ref{fig:entropy_threshold_combo_chart} together show that neither too few nor too many latent steps are desirable. If the latent phase is too short, the model leaves latent exploration before the hidden state is sufficiently organized, so the later explicit CoT cannot fully benefit. If the latent phase is too long, latent computation begins to replace useful explicit verification, which hurts accuracy and can also weaken the overall efficiency--accuracy tradeoff. The key is therefore to find a balanced exit point where latent exploration is sufficient but not excessive. Adaptive switching aims to approximate this balance by using the model's own latent dynamics---including probe entropy and terminating-token probes---to decide when to exit according to the difficulty of the current problem and the model's internal confidence.

\subsection{Failure cases and the limit of hand-crafted switching}
\label{subsec:failure_cases}

The training-free results also reveal an important limitation. On some AIME 2025 problems, latent entropy remains low and the discrete CoT baseline already solves the problem correctly; adding latent steps can still hurt. Low entropy therefore does not by itself imply that the model should remain in, or exit from, the latent phase.
Appendix~\ref{app:entropy_distribution} then analyzes sample-level entropy trajectories. A key finding is that the model does not use the same latent entropy scale for every problem. Instead, it appears to selectively amplify the entropy of the latent phase depending on problem difficulty and internal confidence. By contrast, in ordinary CoT decoding, the peak entropy and overall entropy range are much more similar across samples. In the latent phase, however, different samples can show more than an order-of-magnitude difference in entropy scale.

This failure mode has two implications. First, a single global entropy threshold is too coarse: different problems can require different latent horizons even when their entropy values appear similar. Second, the model should not only be monitored during latent reasoning; it should learn when to stop. The training-free setting establishes that useful switching signals exist, but it also shows that hand-crafted rules only approximate the ideal decision boundary. This motivates the supervised LaTER training procedure in the next section.

\section{Training LaTER}
\label{sec:training_protocol}

The training-free study shows that pretrained models can use latent rollouts, but it also shows that hand-crafted switching is limited. We therefore train a Qwen3-14B model to use a latent segment before explicit reasoning. The trained system differs from the training-free version in two ways: it replaces the pseudo-inverse mapping with a learned projector, and it receives supervision that teaches the model how long the latent segment should be.

\subsection{Model architecture}
\label{sec:model_architecture}

We extend the tokenizer with two boundary tokens, \texttt{<latent\_think>} and \texttt{</latent\_think>}. The embedding layer, transformer backbone, and language-modeling head keep their original architecture, and we add a lightweight projector $g_\phi$ that maps decoder hidden states back into the token embedding space. During latent reasoning, the model computes
\begin{equation}
  h_t = f_\theta(e_t, \mathcal{C}_{<t}), \qquad
  e_{t+1} = g_\phi(h_t),
\end{equation}
where $f_\theta$ is the transformer, $e_t$ is the current latent input embedding, and $\mathcal{C}_{<t}$ is the causal context. This recurrence updates the model's internal reasoning state without emitting a visible token at each latent position.
The supervised assistant format is:
\begin{equation}
  \texttt{<latent\_think>}~l_1,\ldots,l_m~\texttt{</latent\_think>}
  ~\texttt{<think>}~t_1,\ldots,t_n~\texttt{</think>}~a,
\end{equation}
where $l_i$ are latent placeholder positions, $t_i$ are distilled explicit CoT tokens, and $a$ is the final answer. The latent placeholders are not supervised with ordinary token-level cross-entropy (CE). Instead, their input embeddings are replaced by recurrent projector outputs, so gradients from the later explicit reasoning and answer tokens teach the model how to use them as hidden computation steps.

\begin{figure}[h]
  \centering
  \vspace{-1em}
  \includegraphics[width=\linewidth]{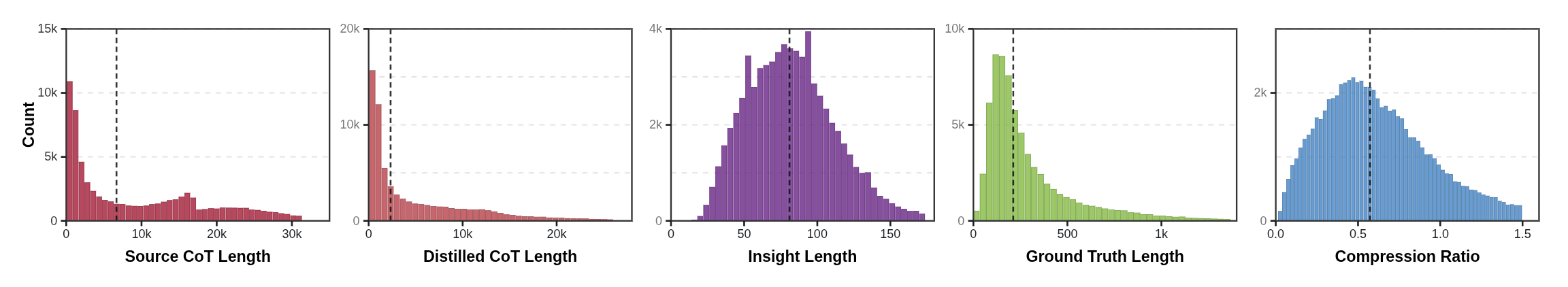}
  \vspace{-2em}
  \caption{Token statistics of the distilled corpus. The figure compares original and distilled reasoning lengths and shows the resulting CoT compression ratios.}
  \label{fig:token_and_compression_distribution_figure}
  \vspace{-1.5em}
\end{figure}

\subsection{Training data construction}
\label{sec:data_construction}

We construct the supervised corpus from reasoning traces sampled from Dolci-Think-SFT-32B~\citep{olmo2026olmo3} and distill them with a stronger reasoning teacher. For each problem, the teacher produces a short \emph{solution intuition}: a few sentences describing the high-level plan without a full derivation. The teacher then generates a shorter explicit CoT conditioned on the original problem and the solution intuition. Each retained record contains a problem, an intuition, a compressed CoT, and a final answer.

The latent budget is tied to the intuition length. If the retained intuition contains $L$ tokens, preprocessing assigns approximately $L/2$ latent steps, subject to maximum-length and tokenization constraints. This design uses the intuition length as a proxy for how much condensed reasoning should be represented by the latent segment.

\begin{wraptable}{r}{0.37\textwidth}
  \centering
  \vspace{-1.5em}
  \small
  \caption{Statistics of the distilled SFT corpus. The compression ratio is the distilled CoT length divided by the original CoT length.}
  \setlength{\tabcolsep}{7pt}
  \renewcommand{\arraystretch}{1.05}
  \begin{tabular}{@{}lcc@{}}
    \toprule
    Statistic & Value & Share \\
    \midrule
    Total examples & 69,745 & 100.0\% \\
    \midrule
    \multicolumn{3}{@{}l}{\textit{Difficulty distribution}} \\
    Easy & 6,667 & 9.5\% \\
    Medium & 45,650 & 65.5\% \\
    Hard & 17,428 & 25.0\% \\
    \midrule
    \multicolumn{3}{@{}l}{\textit{Compression ratio}} \\
    Mean & 0.612 & -- \\
    Median & 0.569 & -- \\
    \midrule
    \multicolumn{3}{@{}l}{\textit{Latent steps}} \\
    Mean & 41.49 & -- \\
    Median & 40.00 & -- \\
    \bottomrule
  \end{tabular}
  \label{tab:data_stats}
  \vspace{-1.5em}
\end{wraptable}

Each training record is rendered as a two-part assistant response. The latent segment contains \texttt{<latent\_think>}, a repeated padding placeholder, and \texttt{</latent\_think>}. The explicit segment contains \texttt{<think>}, the distilled CoT, \texttt{</think>}, and the answer. We also build a teacher-reference conversation in which the problem is paired with the distilled solution intuition and the shortened explicit reasoning trace. This reference provides teacher KL-distribution supervision over explicit reasoning and answer tokens. Figure~\ref{fig:token_and_compression_distribution_figure} summarizes the resulting token counts and the compression ratio between the original and distilled chains of thought, showing that the distilled traces are substantially shorter while still preserving useful reasoning content. Table~\ref{tab:data_stats} gives a compact summary of the dataset. The final training split contains 69,745 examples, with most samples in the medium-difficulty bucket (65.5\%), followed by hard (25.0\%) and easy (9.5\%). The compression ratio has a mean of 0.612 and a median of 0.569, which means that the distilled CoTs keep only about 57--61\% of the original reasoning length. The curriculum metadata groups samples by difficulty so that early training can emphasize easier examples. As detailed in Appendix~\ref{app:latent_sft_data}, the underlying source mix is math- and code-heavy: math accounts for about 37\% of examples and code for about 34\%, while science-oriented questions contribute about 5\% and the remainder mainly comes from instruction-following and general knowledge-oriented prompts.

\subsection{Optimization objective}
\label{sec:optimization_objective}

\begin{figure}[t]
  \centering
  \includegraphics[width=\textwidth]{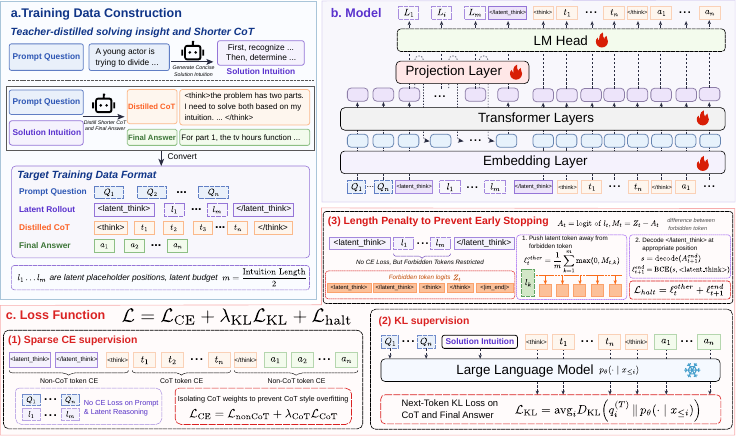}
  \vspace{-1em}
  \caption{\textbf{Overview of training pipeline.} We construct latent-reasoning training sequences with latent and explicit reasoning segments, train the model with supervised and teacher-matching objectives, and obtain a model that first performs latent reasoning and then switches to explicit CoT generation.}
  \label{fig:training_later_overview}
  \vspace{-1em}
\end{figure}

We train with a mixture of supervised language modeling, self-distillation~\cite{zhao2026selfdistilledreasoneronpolicyselfdistillation}, and latent halting supervision. Let $\mathcal{S}_{\mathrm{CoT}}$ denote the interior explicit reasoning positions between \texttt{<think>} and \texttt{</think>}, and let $\mathcal{S}_{\mathrm{nonCoT}}$ denote the remaining supervised response positions, including structural tags and answer tokens. For target token $y_i$, the cross-entropy loss is
\begin{equation}
  \mathcal{L}_{\mathrm{CE}}
  =
  \frac{1}{|\mathcal{S}_{\mathrm{nonCoT}}|}
  \sum_{i \in \mathcal{S}_{\mathrm{nonCoT}}}
  -\log p_\theta(y_i \mid x_{<i})
  +
  \lambda_{\mathrm{CoT}}
  \frac{1}{|\mathcal{S}_{\mathrm{CoT}}|}
  \sum_{i \in \mathcal{S}_{\mathrm{CoT}}}
  -\log p_\theta(y_i \mid x_{<i})
\end{equation}
Only the interior reasoning tokens belong to $\mathcal{S}_{\mathrm{CoT}}$. The boundary tags, latent boundary tokens, answer tokens, and \texttt{<|im\_end|>} belong to $\mathcal{S}_{\mathrm{nonCoT}}$. We set $\lambda_{\mathrm{CoT}}=0.5$ so that explicit reasoning supervision remains useful without overwhelming the structural and answer tokens.

For self-distillation, we precompute top-$k$ teacher distributions with $k=128$. The teacher is the same Qwen3-14B model used to initialize training, but queried with a different input format: we concatenate the original question prompt and the distilled solution intuition into the user message, then feed the distilled short CoT as the assistant continuation and record the teacher distribution at each token position in that short CoT. On these valid teacher positions, the student minimizes a temperature-scaled KL divergence, with temperature $T=1.0$ and weight $\lambda_{\mathrm{KL}}=0.25$.
\begin{equation}
  \mathcal{L}_{\mathrm{KL}}
  =
  \frac{1}{|\mathcal{S}_{\mathrm{KL}}|}
  \sum_{i \in \mathcal{S}_{\mathrm{KL}}}
  D_{\mathrm{KL}}\!\left(q_i^{(T)} \,\|\, p_\theta^{(T)}(\cdot \mid x_{<i})\right)
\end{equation}

Finally, we train the model to terminate latent reasoning at the intended boundary. Let $\mathcal{S}_{\mathrm{lat}}^{\mathrm{int}}$ denote latent interior positions, $\mathcal{S}_{\mathrm{lat}}$ denote all latent positions, $\mathcal{V}_{\mathrm{forbid}}$ be the set of forbidden structural tokens, and $b_i \in \{0,1\}$ indicate whether $i$ is the correct stopping boundary. The raw halt loss is
\begin{equation}
  \mathcal{L}_{\mathrm{halt}}
  =
  \frac{1}{|\mathcal{S}_{\mathrm{lat}}^{\mathrm{int}}|}
  \sum_{i \in \mathcal{S}_{\mathrm{lat}}^{\mathrm{int}}}
  \sum_{v \in \mathcal{V}_{\mathrm{forbid}}}
  \left[z_{i,v} - z_{i,\max}\right]_+
  +
  \frac{1}{|\mathcal{S}_{\mathrm{lat}}|}
  \sum_{i \in \mathcal{S}_{\mathrm{lat}}}
  \mathrm{BCE}\!\left(\sigma(z_{i,\texttt{</latent\_think>}}), b_i\right),
\end{equation}
where $z_{i,v}$ is the logit of token $v$ at position $i$, $z_{i,\max}$ is the largest logit among allowed non-structural tokens, and $\sigma(\cdot)$ is the sigmoid function. The first term penalizes forbidden structural tokens when they become too competitive before the stopping point, while the second term directly trains the model to emit \texttt{</latent\_think>} exactly at the correct boundary. The raw halt loss is assigned a base weight $\lambda_{\mathrm{halt}}^{(s)}=0.025$. To reduce interference between stopping supervision and the main language-modeling objective, we further modulate this term with a dynamic gate,
\begin{equation}
  \alpha_t =
  \mathrm{clip}\!\left(
    \frac{\mathrm{EMA}(\mathcal{L}_{\mathrm{CE}})_t}
    {\mathcal{L}_{\mathrm{CE},t} + \epsilon},
  0, 1\right),
  \qquad
  \mathcal{L}_{\mathrm{halt}}^{\mathrm{eff}}
  =
  \alpha_t \, \lambda_{\mathrm{halt}}^{(s)} \, \mathcal{L}_{\mathrm{halt}}
\end{equation}
where $\mathrm{EMA}(\mathcal{L}_{\mathrm{CE}})_t$ denotes the exponential moving average of the CE loss up to optimization step $t$, $\mathcal{L}_{\mathrm{CE},t}$ is the current-step CE loss, $\epsilon$ is a small constant for numerical stability, and $\mathrm{clip}(\cdot,0,1)$ truncates the gate to the interval $[0,1]$.
The effective halt loss $\mathcal{L}_{\mathrm{halt}}^{\mathrm{eff}}$ therefore applies stopping supervision mainly when it does not conflict with learning the token-level prediction objective.
The total training loss is
\begin{equation}
  \mathcal{L}
  =
  \mathcal{L}_{\mathrm{CE}}
  +
  \lambda_{\mathrm{KL}}\mathcal{L}_{\mathrm{KL}}
  +
  \mathcal{L}_{\mathrm{halt}}^{\mathrm{eff}}
\end{equation}

\subsection{Training setup}
\label{sec:training_setup}

We use AdamW~\citep{loshchilov2018decoupled} with learning rate $1.0\times10^{-7}$, minimum cosine learning rate $1.0\times10^{-8}$, weight decay 0.01, and $\beta_1=0.9$, $\beta_2=0.95$. We enable FlashAttention~2~\citep{dao2023flashattention2fasterattentionbetter}. Distributed training uses DeepSpeed ZeRO-3~\citep{rajbhandari2020zero}, and the launcher uses 8*Nvidia A800 80G GPUs on a single node.

For evaluation, we compare four systems on the same benchmark: the original Qwen3-14B with standard explicit CoT prompting, a CoT SFT model trained on the same distilled data using only the CE and KL objectives, LaTER in the training-free setting, and the fully trained LaTER model. The CoT SFT baseline uses exactly the same training data and optimization strategy as trained LaTER, but removes the latent-reasoning component and therefore does not include the $\mathcal{L}_{\mathrm{halt}}^{\mathrm{eff}}$.

\definecolor{trainaccbest}{RGB}{214,239,223}
\definecolor{traintokbest}{RGB}{221,233,248}
\newcommand{\trainbestacc}[2]{\cellcolor{trainaccbest}\textbf{#1} {\scriptsize(+#2)}}
\newcommand{\trainbesttok}[2]{\cellcolor{traintokbest}\textbf{#1} {\scriptsize($\downarrow$#2\%)}}
\newcommand{\trainaccgain}[2]{#1 {\scriptsize(+#2)}}
\newcommand{\traintokgain}[2]{#1 {\scriptsize($\downarrow$#2\%)}}
\newcommand{\trainlater}{\textbf{LaTER}}
\begin{table}[t]
  \centering
  \small
  \setlength{\tabcolsep}{3.2pt}
  \renewcommand{\arraystretch}{1.08}
  \caption{Comparison on the benchmarks between different baselines. Green cells mark the best accuracy among all methods for each benchmark, while blue cells mark the lowest token usage.}
  \label{tab:training_bench}
  \resizebox{\columnwidth}{!}{%
    \begin{tabular}{@{}l@{\hspace{4pt}}cc@{\hspace{5pt}}cc@{\hspace{5pt}}cc@{\hspace{5pt}}cc@{\hspace{5pt}}cc@{\hspace{5pt}}cc@{\hspace{5pt}}cc@{}}
      \toprule
      \multirow{2}{*}{Method}
      & \multicolumn{2}{c}{AIME 2025}
      & \multicolumn{2}{c}{GSM8K}
      & \multicolumn{2}{c}{MATH-500}
      & \multicolumn{2}{c}{ARC-Challenge}
      & \multicolumn{2}{c}{GPQA}
      & \multicolumn{2}{c}{HumanEval+}
      & \multicolumn{2}{c}{MBPP+} \\
      \cmidrule(lr){2-3}\cmidrule(lr){4-5}\cmidrule(lr){6-7}\cmidrule(lr){8-9}\cmidrule(lr){10-11}\cmidrule(lr){12-13}\cmidrule(lr){14-15}
      & Acc. & Token & Acc. & Token & Acc. & Token & Acc. & Token & Acc. & Token & Acc. & Token & Acc. & Token \\
      \midrule
      CoT Baseline & 70.0 & 15730 & 92.7 & 879 & 93.4 & 3472 & 96.3 & 631 & 62.6 & 6301 & 90.8 & 2427 & 82.0 & 2144 \\
      CoT-SFT & 73.3 & 12687 & 94.1 & 743 & 94.0 & 3014 & 95.5 & 578 & 61.1 & 5885 & 90.2 & 2071 & 78.0 & 1817 \\
      \midrule
      \begin{tabular}[c]{@{}l@{}}\trainlater\\ (training-free)
      \end{tabular}
      & \trainaccgain{73.3}{3.3} & \traintokgain{10661}{32}
      & \trainaccgain{93.2}{0.5} & \traintokgain{649}{26}
      & \trainbestacc{97.2}{3.8} & \traintokgain{2887}{17}
      & \trainaccgain{96.5}{0.2} & \traintokgain{533}{16}
      & 60.6 & \traintokgain{4445}{29}
      & 90.8 & \traintokgain{1790}{26}
      & 79.6 & \traintokgain{1760}{18} \\
      \begin{tabular}[c]{@{}l@{}}\trainlater\\ (training)
      \end{tabular}
      & \trainbestacc{80.0}{10.0} & \trainbesttok{10575}{33}
      & \trainbestacc{94.8}{2.1} & \trainbesttok{644}{27}
      & \trainaccgain{96.4}{3.0} & \trainbesttok{2541}{27}
      & \trainbestacc{96.7}{0.4} & \trainbesttok{434}{31}
      & \trainbestacc{63.1}{0.5} & \trainbesttok{3885}{38}
      & \trainbestacc{92.2}{1.4} & \trainbesttok{1776}{27}
      & \trainbestacc{82.8}{0.8} & \trainbesttok{1717}{20} \\
      \bottomrule
    \end{tabular}%
  }
  \vspace{-1em}
\end{table}

\subsection{Main results}
\label{sec:training_results}

Table~\ref{tab:training_bench} compares the four Qwen3-14B variants. Trained LaTER achieves the lowest token usage on all seven benchmarks and the best accuracy on most of them. On AIME 2025, it reaches 80.0\% accuracy, 10.0 points above the standard CoT baseline, while reducing average token usage by 33\%. It also improves GSM8K, ARC-Challenge, GPQA, HumanEval+, and MBPP+ relative to the baseline while using fewer tokens.

\textbf{Transcending the Accuracy-Efficiency Trade-Off.} A key comparison is CoT-SFT versus trained LaTER. CoT-SFT benefits from the same distilled data and improves AIME 2025 from 70.0\% to 73.3\%, but it remains less accurate than trained LaTER and uses more tokens (12,687 versus 10,575 on AIME 2025). This suggests that the gains are not merely a consequence of shorter supervised traces: the latent-first architecture contributes additional efficiency and reasoning accuracy.

\textbf{Isolating the Role of Latent Reasoning.} The results are also nuanced. Training-free LaTER remains the strongest method on MATH-500 in Table~\ref{tab:training_bench}, whereas trained LaTER is more efficient and stronger on most other tasks. This indicates that supervised latent training improves the overall accuracy--efficiency frontier but does not uniformly dominate every benchmark. We view this as evidence that latent-budget allocation and data mixture remain important design choices.

\section{Related Works}
\label{sec:related_works}

\textbf{Training-free latent reasoning.}
Soft Thinking and SwiReasoning~\citep{zhang2026softthinking, shi2026swireasoningswitchthinkinglatentexplicit} replace hard token inputs with probability-weighted mixtures of token embeddings, enabling latent reasoning from the model's own next-token distribution. However, soft-embedding methods can collapse toward the dominant token and thus behave similarly to greedy decoding, limiting their ability to maintain alternative reasoning paths. SeLaR~\citep{fu2026selarselectivelatentreasoning} addresses this issue with entropy-gated activation, applying latent reasoning only at high-uncertainty steps and preserving discrete decoding at deterministic steps. LatentMAS does not use next token embedding mixtures~\citep{zou2025latentcollaborationmultiagentsystems}. Instead, it projects the previous step's hidden state back into the input embedding space and uses this latent state as the next input. It further shares KV-cache working memory across agents as a training-free communication channel. While effective, these training-free methods still rely on hand-crafted switching or local confidence heuristics and do not explicitly separate exploratory reasoning from rigorous derivation.

\textbf{Training-based latent reasoning.}
Coconut~\citep{hao2025training} pioneers autoregressive latent reasoning by feeding the last-layer hidden state back as the next input embedding, showing that continuous thoughts can support implicit breadth-first exploration. However, its reliance on fixed latent steps and direct hidden-state reuse exposes a mismatch between hidden states and token embeddings. Subsequent methods improve this paradigm by learning better latent interfaces. SoftCoT~\citep{xu2025softcot} reduces full-model adaptation by using an assistant model to generate soft thoughts and a trainable projection module to align them with the target LLM. More recent methods further refine the definition and training of latent tokens. Latent-SFT~\citep{deng2025latent} constrains latent reasoning to the vocabulary column space and learns latent tokens with KL and CE objectives, whereas CoLaR~\citep{tan2026think} predicts compressed embedding distributions and applies reinforcement learning to encourage both diverse exploration and compact reasoning. These methods demonstrate that latent reasoning can substantially shorten reasoning chains, but they largely aim to replace explicit CoT with latent computation. As a result, their performance can degrade on complex tasks where precise symbolic verification is essential.

\section{Conclusion}
\label{sec:conclusion}

We introduce LaTER, a latent-then-explicit reasoning paradigm for reducing test-time token cost without discarding explicit verification. The method separates reasoning into two phases: a continuous latent rollout for early exploration, followed by discrete CoT generation for symbolic checking and final-answer construction. In the training-free setting, we have found that pretrained reasoning models already exhibit structured latent trajectories, including terminating-token probes and informative entropy dynamics. A simple adaptive switch based on these signals reduces token usage and can improve accuracy on several benchmarks. We then construct \textsc{Latent-Switch-69K} and train a LaTER model with a learned latent projector and halting supervision. The trained model improves the accuracy--efficiency tradeoff across mathematics, coding, and knowledge-intensive benchmarks, reaching 80.0\% on AIME 2025 while using one third fewer tokens than standard CoT.

LaTER also leaves open important questions. The training-free switch is still a hand-crafted approximation, and the trained model's behavior depends on the latent-budget distribution and the quality of distilled intuitions. Future work should learn richer instance-adaptive halting policies, study longer latent exploration for open-ended tasks, and extend latent-then-explicit reasoning to multimodal settings where full verbalization can be especially costly.




\medskip

\bibliographystyle{unsrtnat}
\bibliography{main}


\appendix

\clearpage

\section{Per-step entropy heterogeneity in training-free latent reasoning}
\label{app:entropy_distribution}

We provide a finer-grained view of the stopping signals used by the training-free version of LaTER.
For each AIME 2025 problem, we begin with latent reasoning and continue rolling the hidden state forward until the current hidden state can be decoded to a model-native terminating token, such as \texttt{<|im\_end|>}.
At every latent step, we decode the hidden state into the vocabulary space and record the entropy of the resulting predictive distribution.
This produces a trajectory-level entropy trace for the entire latent phase, from the first latent transition to the step immediately preceding termination.

\begin{figure}[h]
  \centering
  \includegraphics[width=0.9\linewidth]{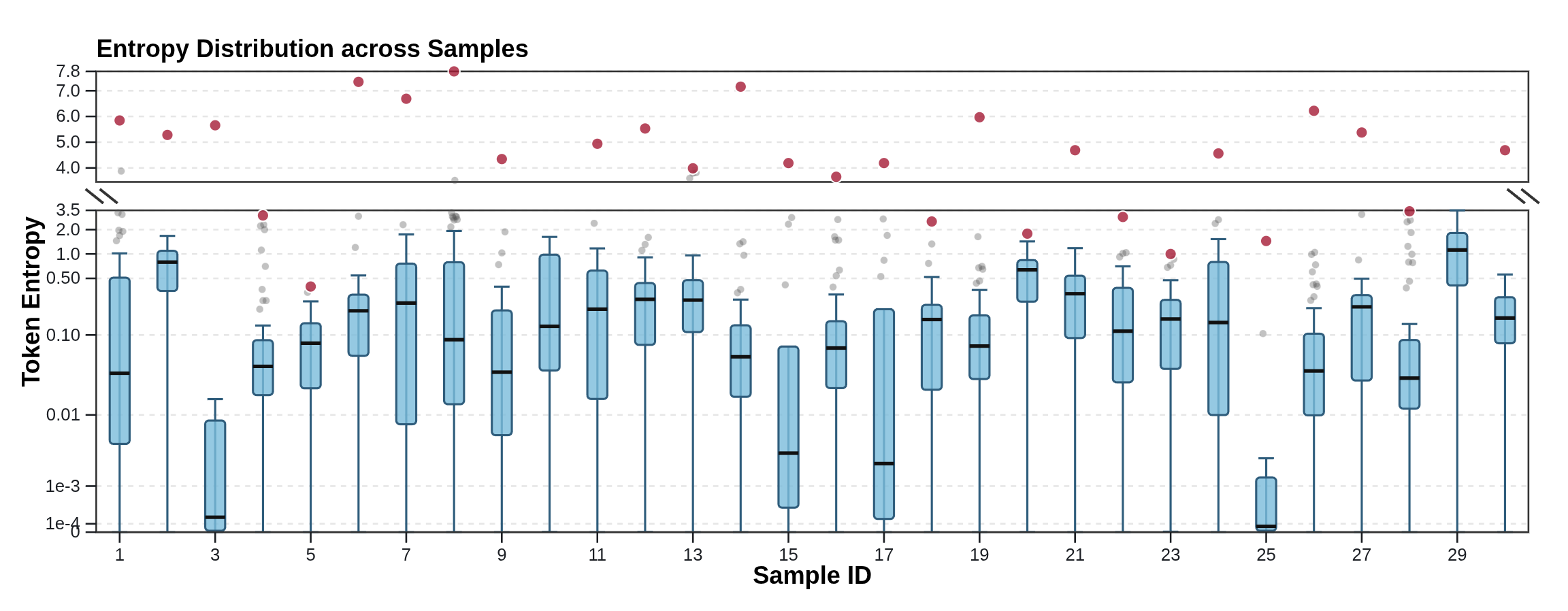}
  \caption{Per-step entropy distributions during the training-free latent phase on AIME 2025.
    For each problem, we record the entropy of the decoded vocabulary distribution at every latent step until the hidden state decodes to a terminating token such as \texttt{<|im\_end|>}.
    Each boxplot aggregates all problems that are still in the latent phase at that step position.
  The wide variation in spread and upper tails shows that both the scale and the timing of peak entropy differ substantially across instances, which helps explain why a single entropy threshold cannot be uniformly optimal.}
  \label{fig:latent_token_entropy_boxplot}
\end{figure}

Figure~\ref{fig:latent_token_entropy_boxplot} groups the entropy traces by latent step position.
Each boxplot shows the entropy values at one step, using all problems that are still in the latent phase at that point.
The distributions vary a lot across steps and across problems.
At many step positions, the spread is wide, the upper tail changes noticeably, and the number of active trajectories also changes as some problems terminate earlier than others.
This means that the latent phase does not follow one shared entropy curve.

The main finding is that both the size of the entropy peak and the step at which it appears differ greatly across problems.
Some problems show a sharp early peak and then settle quickly.
Others stay at relatively low entropy for many steps and peak only near the stopping point.
Still others remain broad and unstable until very late in the trajectory, suggesting that the model continues exploring for much longer.
In short, the entropy maximum is highly instance-specific in both value and timing.

This is why a single global threshold is only a rough stopping rule.
A threshold that works well for trajectories with large early spikes may stop too early on problems that need a longer latent phase.
A higher threshold may fit those slower cases better, but then it may wait too long on problems that are already ready to switch.
In practice, the decision should depend not only on the entropy at one step, but also on the trend of the trajectory: whether entropy is rising or falling, how long uncertainty lasts, and whether the decoded state is already close to a terminating token.

For this reason, we view the training-free entropy rule as a useful diagnostic rather than a complete solution.
It already captures meaningful structure in the latent dynamics and yields strong efficiency gains in the main experiments.
However, the appendix results also show that a hand-crafted threshold cannot fully match the diversity of real latent trajectories.
A more natural next step is to learn an instance-adaptive switching policy that uses the full trajectory, instead of relying on one fixed scalar cutoff.

\section{Construction of Latent-Switch-69K}
\label{app:latent_sft_data}

This section describes how we build the supervised corpus used to train LaTER.
Each retained example contains four parts: a user problem, a distilled solution intuition, a shortened explicit CoT, and a final answer.
The preprocessing pipeline turns this record into a latent-supervised fine-tuning example.
In this format, the model first passes through a bounded latent segment and then returns to ordinary explicit reasoning.
The same pipeline also builds a teacher-reference conversation.
This allows us to combine token-level language modeling targets with teacher-distribution supervision on explicit reasoning and answer tokens.

\begin{figure}[h]
  \centering
  \includegraphics[width=0.9\linewidth]{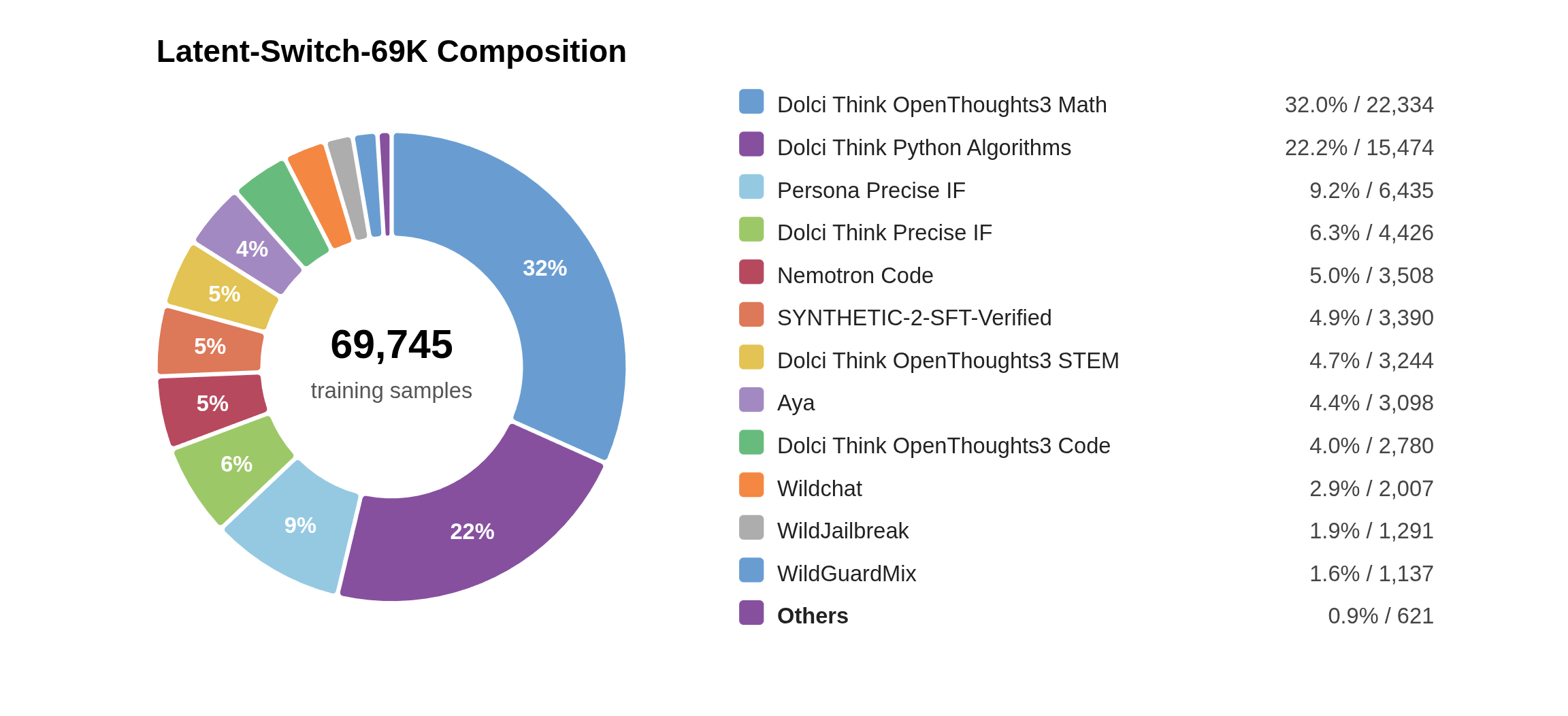}
  \caption{Composition of the Latent-Switch-69K training corpus after filtering and distillation.}
  \label{fig:training_dataset_composition_pie}
\end{figure}

Figure~\ref{fig:training_dataset_composition_pie} summarizes the composition of the final corpus.
Consistent with Table~\ref{tab:data_stats}, the final training split contains 69{,}745 examples.
Most examples are in the medium-difficulty bucket, which accounts for 65.5\% of the data.
Hard examples account for 25.0\%, and easy examples account for 9.5\%.
At the domain level, the source mixture is dominated by mathematical and coding data: math contributes about 37\% of examples and code about 34\%, while science-oriented questions account for roughly 5\%.
The remaining examples mainly come from instruction-following and general knowledge-oriented prompts, so the retained corpus stays centered on reasoning-intensive tasks while preserving some diversity in format and topic.
This imbalance is intentional.
Medium-difficulty problems provide the cleanest signal for learning when to move from latent exploration to explicit verification.
They are hard enough to require real reasoning, but usually not so noisy that distillation becomes unstable.
The hard subset broadens coverage and exposes the model to longer reasoning chains.
The easy subset helps stabilize training and preserves short-form answer behavior.

\paragraph{Distillation pipeline.}
We start from reasoning traces sampled from Dolci-Think-SFT-32B.
We then distill them with a stronger reasoning model.
For each source problem, we first ask for a short \emph{solution intuition} that states the high-level plan in a few sentences.
We then ask the teacher to produce a shorter explicit CoT conditioned on the original problem and this intuition.
The final record therefore keeps both a compressed latent-style summary and an explicit derivation.
This design lets us train the model to use continuous latent computation without giving up token-level supervision on the visible reasoning segment.

\paragraph{Student response template.}
For a sample with $m$ latent steps, distilled CoT tokens $t_{1:n}$, and answer tokens $a_{1:r}$, the assistant response is written as
$$
\texttt{<latent\_think>}
~l_1,\ldots,l_m~
\texttt{</latent\_think>}
~\texttt{<think>}
~t_1,\ldots,t_n~
\texttt{</think>}
~a_1,\ldots, ~a_r
\texttt{<|im\_end|>}.
$$
The symbols $l_1,\ldots,l_m$ denote latent placeholder positions.
In the current implementation, these positions are filled with a repeated placeholder token.
However, they are not trained as ordinary language targets.
During the forward pass, their token embeddings are replaced by recurrent latent states produced by the latent projector.
The model therefore learns to reason internally across these positions instead of emitting visible text there.

\paragraph{Latent budget assignment.}
Each sample stores its latent budget as an integer field, which we denote conceptually as \texttt{n\_latent\_steps}.
This value determines how many placeholder positions appear between \texttt{<latent\_think>} and \texttt{</latent\_think>}.
In preprocessing, the budget is derived from the length of the distilled solution intuition.
If the retained intuition contains $L$ tokens, then the target latent budget is set to about $L/2$.
This value is then clipped by the maximum latent length and other tokenization constraints.
This heuristic ties the amount of latent computation to the amount of compressed reasoning content kept by distillation.
Across the final corpus, the latent-step count has mean 41.49 and median 40.00.
This choice is determined by experimental phenomena in the training-free setting. In section~\ref{subsec:adaptive_results_training_free}, we found that the model naturally has the best reasoning accuracy and token efficiency after executing about 40-50 steps of latent reasoning, so we normalized the number of latent steps in the training data to this range.

\paragraph{Supervision masks.}
Let $\mathcal{S}_{\mathrm{prompt}}$ denote prompt positions.
Let $\mathcal{S}_{\mathrm{lat}}^{\mathrm{int}}$ denote latent interior positions, excluding the latent boundary markers.
The CE label at token position $i$ is
$$
y_i=
\left\{
  \begin{array}{ll}
    -100, & i \in \mathcal{S}_{\mathrm{prompt}} \cup \mathcal{S}_{\mathrm{lat}}^{\mathrm{int}},\\
    x_i, & \mbox{otherwise}.
  \end{array}
  \right.
  $$
  Prompt tokens and latent interior placeholders are therefore masked from ordinary token-level CE.
  By contrast, the latent boundary tokens, explicit CoT tokens, answer tokens, and the terminal \texttt{<|im\_end|>} token remain supervised.
  The preprocessing stage also builds dedicated masks for latent boundaries, explicit CoT regions, answer regions, and teacher-KL positions.
  This ensures that each objective is applied only where it is semantically meaningful.

  \paragraph{Teacher reference construction.}
  For teacher-distribution supervision, each example also includes a teacher-reference conversation.
  This reference omits the student's latent placeholder segment and begins directly with the explicit reasoning part,
  $$
  \texttt{<think>}~t_1,\ldots,t_n~\texttt{</think>}~a_1,\ldots,a_r.
  $$
  Operationally, the teacher input pairs the original question with the distilled solution intuition.
  The shortened explicit CoT and final answer are then treated as the continuation to be matched.
  This provides teacher hidden states and teacher logits that align with explicit reasoning and answer positions.
  It does not require the teacher to model the student's continuous latent placeholders.
  Teacher KL is applied only on positions selected by the teacher-KL mask.

  Finally, the filtered corpus preserves the compression effect that motivates LaTER.
  As reported in Table~\ref{tab:data_stats}, the distilled CoT compression ratio has mean 0.612 and median 0.569.
  The visible reasoning trace is therefore typically only about 57--61\% as long as the original one.
  The dataset does not merely shorten responses.
  It explicitly separates condensed latent planning from explicit symbolic verification.
  That structural decomposition is what makes the later latent-reasoning finetuning objective well posed.

\section{Training Details}
\label{app:training_details}

We initialize LaTER from a Qwen3-14B backbone and optimize the model end to end so that it can interleave a continuous latent reasoning segment with an explicit textual reasoning segment. Each assistant response is formatted as
\[
  \texttt{<latent\_think>}
  ~l_1,\ldots,l_m~
  \texttt{</latent\_think>}
  ~
  \texttt{<think>}
  ~t_1,\ldots,t_n~
  \texttt{</think>}
  ~a~
  \texttt{<|im\_end|>}.
\]
Here $l_{1:m}$ denote latent placeholder positions, $t_{1:n}$ denote explicit CoT tokens, and $a$ denotes the final answer. During training, the latent placeholders are not treated as ordinary language targets. Instead, their token embeddings are replaced by recurrent latent states produced by a learned latent projector.

\paragraph{Latent forward pass.}
For each example, the model first processes the prompt and latent prefix with a cache-based recurrent rollout. At latent step $t$, the previous hidden state $h_{t-1}$ is mapped by the latent projector to the next latent embedding,
\[
  e^{\mathrm{lat}}_t = g_{\phi}(h_{t-1}).
\]
These projected embeddings are then written back into the full input embedding sequence. The model subsequently performs a causal teacher-forcing forward pass over the entire sequence, using ordinary token embeddings outside the latent interior and projected latent embeddings inside the latent segment. During batch construction, we align both the assistant prefix and the transition into \texttt{<think>} across examples, which keeps the latent-to-text boundary synchronized during distributed training.

\paragraph{Supervised objectives.}
Prompt tokens and latent interior placeholders are masked from token-level cross-entropy. The latent boundary tokens, explicit reasoning tokens, answer tokens, and the final \texttt{<|im\_end|>} token remain supervised. We decompose the CE objective into an explicit-CoT term and a complementary non-CoT term,
\[
  \mathcal{L}_{\mathrm{CE}}
  =
  \mathcal{L}_{\mathrm{nonCoT}} +
  \lambda_{\mathrm{CoT}}\mathcal{L}_{\mathrm{CoT}}.
\]
In the current configuration, $\lambda_{\mathrm{CoT}}=0.5$, so explicit reasoning tokens and the remaining supervised tokens contribute at the similar scale. The structural tags \texttt{<think>} and \texttt{</think>} are assigned to $\mathcal{L}_{\mathrm{nonCoT}}$, whereas $\mathcal{L}_{\mathrm{CoT}}$ contains only the interior reasoning tokens.

\paragraph{Teacher distribution matching.}
In addition to CE, we apply cached teacher-distribution supervision on selected explicit reasoning and answer positions. Let $q_i$ denote the cached top-$k$ teacher distribution and let $p_{\theta}(\cdot\mid i)$ denote the student distribution at the aligned source position. The KL objective is
\[
  \mathcal{L}_{\mathrm{KL}}
  =
  \frac{1}{|\mathcal{S}_{\mathrm{KL}}|}
  \sum_{i \in \mathcal{S}_{\mathrm{KL}}}
  D_{\mathrm{KL}}\!\left(q_i \,\|\, p_{\theta}(\cdot\mid i)\right).
\]
The current configuration uses temperature $1.0$ and KL weight $0.25$. This objective distills the teacher's explicit reasoning and answer behavior without requiring the teacher to model the student's continuous latent placeholders.

Importantly, we do not apply CE or KL supervision directly inside the latent reasoning segment. The latent placeholder positions are not trained to match token targets or teacher distributions. Instead, the latent segment is optimized only indirectly: gradients from the downstream explicit CoT and answer tokens are back-propagated through the latent rollout. In this way, the model learns latent reasoning states only to the extent that they help the later explicit reasoning and final answer.

\paragraph{Halting supervision.}
We further train the latent segment to terminate at the correct boundary with a dense auxiliary halting loss over latent interior positions. This loss compares the logit of \texttt{</latent\_think>} and other forbidden structural tokens against the best allowed non-structural token, while also applying a BCE loss that toward the correct latent boundary. The halting weight is annealed to a small final value of $0.025$ and gated by the CE quality signal so that the stopping objective does not dominate the language-modeling objective.

\paragraph{Optimization setup.}
The current training run uses bf16 training, FlashAttention-2, gradient checkpointing, DeepSpeed ZeRO-3, micro-batch size $1$, and gradient accumulation $4$. We set the maximum sequence length to $24{,}096$, allow latent budgets of up to $128$ steps.

\paragraph{Compute resources.}
All training runs are conducted on Nvidia A800 80GB GPUs. The trained LaTER model is optimized on a single node with 8 GPUs and requires approximately 5 days for the main training run under the configuration above. The CoT-SFT baseline is trained on the same hardware setup and requires approximately 2 days. All evaluation tasks are run on 2 A800 80GB GPUs. These numbers are intended to give a practical estimate of the wall-clock compute required to reproduce the reported training and evaluation pipeline.

\paragraph{Evaluation protocol.}
Unless otherwise specified, each reported evaluation result is obtained by running the corresponding model-setting pair multiple times under the same decoding configuration with a fixed random seed.

\clearpage
\section{Prompts}
\label{app:prompts}

\begingroup
\raggedbottom
\makeatother
\tcbset{promptbox/.style={
    colback=white,
    colframe=black,
    fontupper=\ttfamily\footnotesize,
}}

\begin{figure}[h]
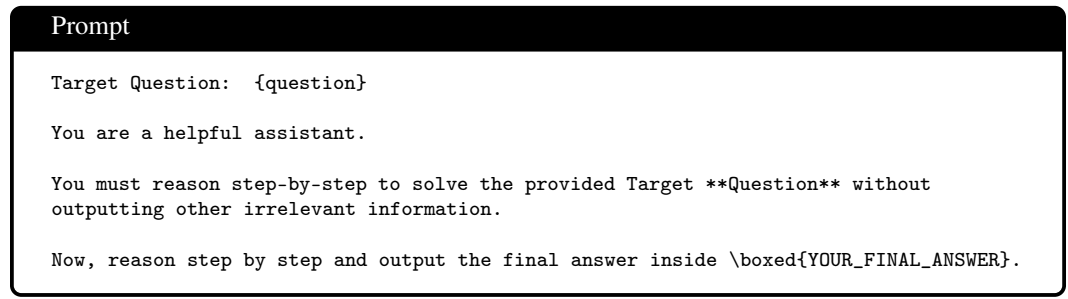

  \begin{tcolorbox}[
      promptbox,
      title=Prompt
    ]
    Target Question: \{question\}\\
    \\
    You are a helpful assistant.\\
    \\
    You must reason step-by-step to solve the provided Target **Question** without outputting other irrelevant information.\\
    \\
    Now, reason step by step and output the final answer inside \textbackslash boxed\{YOUR\_FINAL\_ANSWER\}.
  \end{tcolorbox}
  \caption{Prompt for math questions.}
  \label{fig:prompt_math}
\end{figure}

\begin{figure}[h]
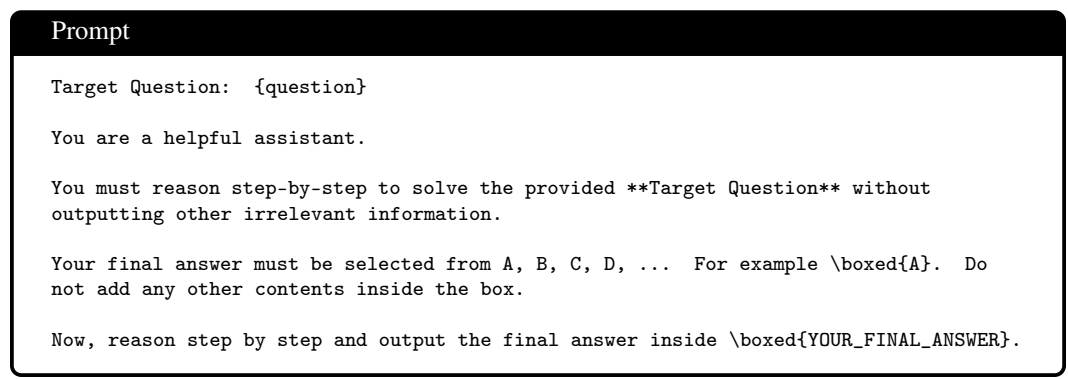

  \begin{tcolorbox}[
      promptbox,
      title=Prompt
    ]
    Target Question: \{question\}\\
    \\
    You are a helpful assistant.\\
    \\
    You must reason step-by-step to solve the provided **Target Question** without outputting other irrelevant information.\\
    \\
    Your final answer must be selected from A, B, C, D, ... For example \textbackslash boxed\{A\}. Do not add any other contents inside the box.\\
    \\
    Now, reason step by step and output the final answer inside \textbackslash boxed\{YOUR\_FINAL\_ANSWER\}.
  \end{tcolorbox}
  \caption{Prompt for multi-choice questions.}
  \label{fig:prompt_choice}
\end{figure}

\begin{figure}[h]
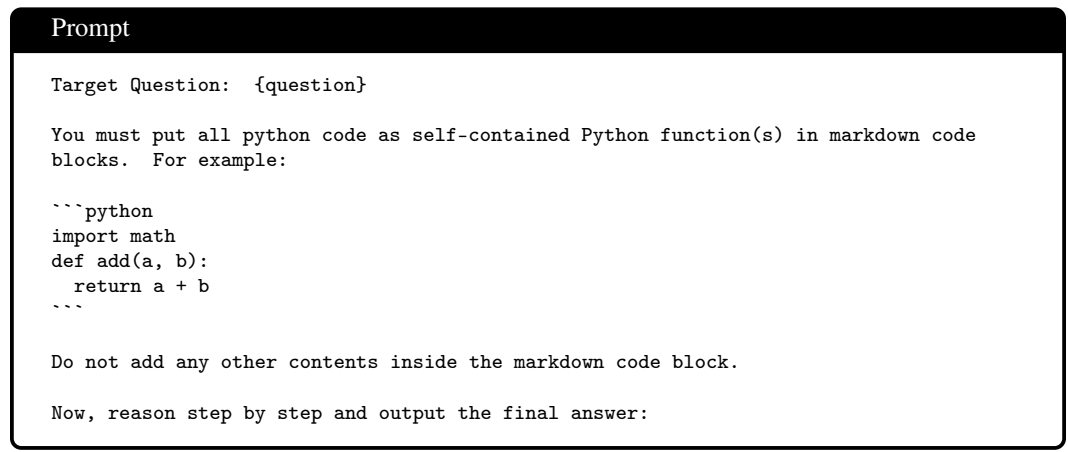

  \begin{tcolorbox}[
      promptbox,
      title=Prompt
    ]
    Target Question: \{question\}\\
    \\
    You must put all python code as self-contained Python function(s) in markdown code blocks. For example:\\
    \\
    \textasciigrave\textasciigrave\textasciigrave{}python\\
    import math\\
    def add(a, b):\\
    \hspace*{1em}return a + b\\
    \textasciigrave\textasciigrave\textasciigrave{}\\
    \\
    Do not add any other contents inside the markdown code block.\\
    \\
    Now, reason step by step and output the final answer:
  \end{tcolorbox}
  \caption{Prompt for code problems.}
  \label{fig:prompt_code}
\end{figure}

\begin{figure}[h]
  \begin{tcolorbox}[
      promptbox,
      title=Prompt
    ]
    You are an expert reasoning data curator.\\
    \\
    Extract only the key insights from the source reasoning.\\
    \\
    Rules:\\
    - Return valid JSON only.\\
    - Do not produce a short chain of thought.\\
    - Do not provide any final answers in your response.\\
    - \texttt{correct\_insight} must be the coarse but correct high-level solution idea.\\
    - \texttt{incorrect\_insights} must only include wrong ideas that are explicitly evidenced in the source outputs.\\
    - Do not invent errors that are not present in the source outputs.
  \end{tcolorbox}
  \caption{System prompt to generate solution intuition.}
  \label{fig:prompt_system_data_construct}
\end{figure}

\begin{figure}[h]
  \begin{tcolorbox}[
      promptbox,
      title=Prompt
    ]
    You will receive a problem, one or more source outputs, and optional ground truth.\\
    \\
    Return JSON with this schema:\\
    \{\\
      \hspace*{1em}"task\_summary": "short task type description",\\
      \hspace*{1em}"correct\_insight": "2--10 sentences of high-level correct plan",\\
      \hspace*{1em}"incorrect\_insights": [\\
        \hspace*{2em}\{\\
          \hspace*{3em}"idea": "2--10 sentences of a wrong high-level plan evidenced in the source outputs",\\
          \hspace*{3em}"why\_wrong": "brief reason it fails"\\
        \hspace*{2em}\}\\
      \hspace*{1em}],\\
      \hspace*{1em}"source\_answer\_correct": true,\\
      \hspace*{1em}"contains\_reflection": true\\
    \}\\
    \\
    REMEMBER: Only extract insights that reflect on the high-level idea. **Do not** give any actual answer in \texttt{correct\_insight} or \texttt{incorrect\_insights}.\\
    \\
    Only give your complete plan to solve the question. Do not directly state whether it is right or wrong. Do not add content unrelated to the idea in \texttt{correct\_insight} or \texttt{incorrect\_insights}.\\
    \\
    Problem prompt:\\
    $<\!\!<\!\!<$PROMPT$>\!\!>\!\!>$\\
    \{prompt\}\\
    $<\!\!<\!\!<$END\_PROMPT$>\!\!>\!\!>$\\
    \\
    Source outputs:\\
    $<\!\!<\!\!<$OUTPUTS$>\!\!>\!\!>$\\
    \{outputs\}\\
    $<\!\!<\!\!<$END\_OUTPUTS$>\!\!>\!\!>$\\
    \\
    Ground truth:\\
    $<\!\!<\!\!<$GROUND\_TRUTH$>\!\!>\!\!>$\\
    \{ground\_truth\}\\
    $<\!\!<\!\!<$END\_GROUND\_TRUTH$>\!\!>\!\!>$\\
    \\
    Metadata:\\
    \{metadata\_json\}
  \end{tcolorbox}
  \caption{User prompt to generate solution intuition.}
  \label{fig:prompt_user_intuition}
\end{figure}

\begin{figure}[h]
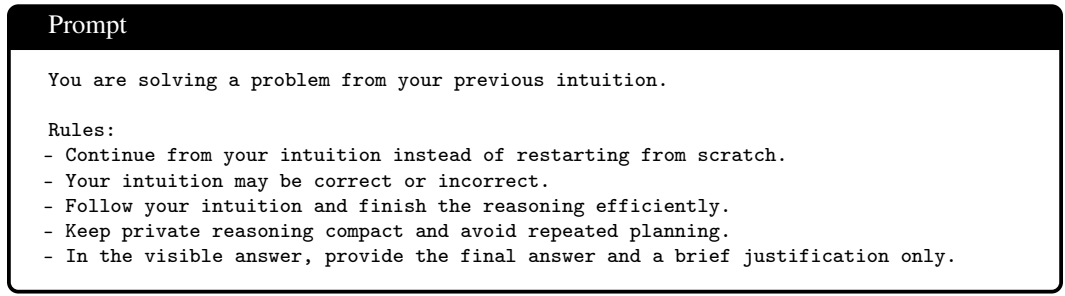

  \begin{tcolorbox}[
      promptbox,
      title=Prompt
    ]
    You are solving a problem from your previous intuition.\\
    \\
    Rules:\\
    - Continue from your intuition instead of restarting from scratch.\\
    - Your intuition may be correct or incorrect.\\
    - Follow your intuition and finish the reasoning efficiently.\\
    - Keep private reasoning compact and avoid repeated planning.\\
    - In the visible answer, provide the final answer and a brief justification only.
  \end{tcolorbox}
  \caption{System prompt to generate shorter CoT.}
  \label{fig:prompt_system_cot}
\end{figure}

\begin{figure}[t]
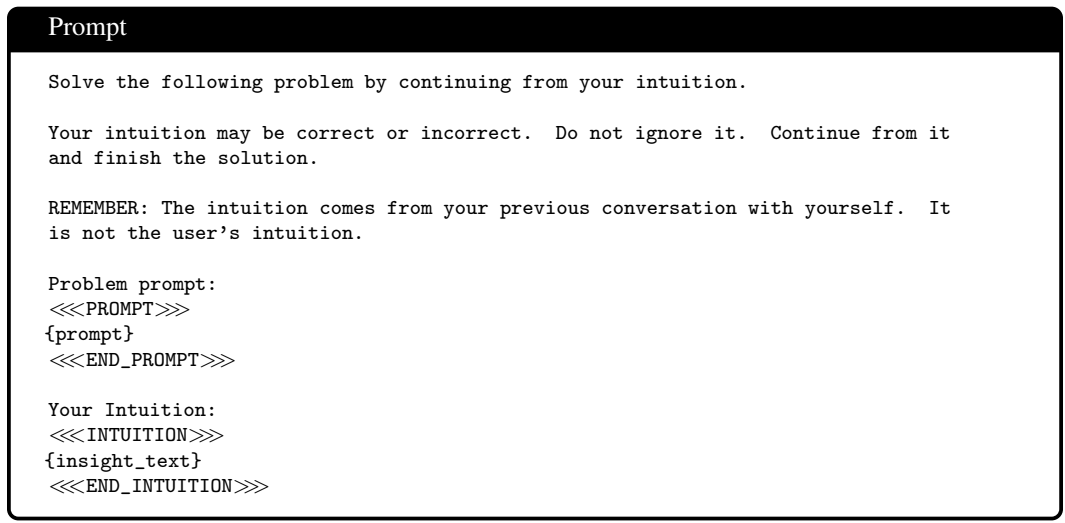

  \begin{tcolorbox}[
      promptbox,
      title=Prompt
    ]
    Solve the following problem by continuing from your intuition.\\
    \\
    Your intuition may be correct or incorrect. Do not ignore it. Continue from it and finish the solution.\\
    \\
    REMEMBER: The intuition comes from your previous conversation with yourself. It is not the user's intuition.\\
    \\
    Problem prompt:\\
    $<\!\!<\!\!<$PROMPT$>\!\!>\!\!>$\\
    \{prompt\}\\
    $<\!\!<\!\!<$END\_PROMPT$>\!\!>\!\!>$\\
    \\
    Your Intuition:\\
    $<\!\!<\!\!<$INTUITION$>\!\!>\!\!>$\\
    \{insight\_text\}\\
    $<\!\!<\!\!<$END\_INTUITION$>\!\!>\!\!>$
  \end{tcolorbox}
  \caption{User prompt to generate shorter CoT.}
  \label{fig:prompt_user_cot}
\end{figure}

\endgroup

\clearpage
\section{Failure Case: Confident but Misleading Latent Reasoning}
\label{app:failure_low_entropy_latent}

\tcbset{promptbox/.style={
    colback=white,
    colframe=black,
}}

Figure~\ref{fig:failure_low_entropy_latent} presents a representative failure case from AIME 2025 Problem~10. The problem asks the model to count valid fillings of a $3\times9$ Sudoku-band grid, where each row and each $3\times3$ block contains the digits $1,\ldots,9$ exactly once. The correct answer is $81$.

This example is notable because the latent trajectory does not appear uncertain under the entropy diagnostic. During latent reasoning, entropy remains low, with maximum entropy only $1.625$. The highest entropy value in the CoT stage is $3.23$. The latent reasoning ends after the trajectory hits \texttt{<|endoftext|>}. The subsequent explicit reasoning phase follows an incorrect counting path and returns $66$.

In contrast, the same backbone under standard explicit CoT decoding solves the problem correctly and returns $81$. The explicit CoT trace preserves the intermediate combinatorial bookkeeping: it fixes the first row, counts the possible second-row block assignments, accounts for the independent within-block permutations in the third row, and obtains $9!\cdot 56\cdot 6^6 = 2^{16}3^{10}5^1 7^2$. The LaTER trace enters the explicit phase after a low-entropy latent segment, reconstructs a similar high-level counting setup, but then treats the final block as determined once the first two blocks are chosen. This omits an additional $6^3$ ordering factor and yields $9!\cdot 56\cdot 6^3 = 2^{13}3^7 5^1 7^2$, producing the wrong answer $66$.

This case illustrates a limitation of entropy as a standalone confidence signal for latent reasoning. Low entropy can indicate that the model is locally confident, but it does not guarantee that the latent state encodes the correct global combinatorial structure. Here, latent reasoning appears to compress the search into a confident but flawed state, while explicit CoT preserves enough intermediate structure to recover the missing factor.

\begin{figure}[t]
  \centering
  \begin{tcolorbox}[
      promptbox,
      title=Failure Case: AIME 2025 Problem 10
    ]
    \textbf{Task.}
    The 27 cells of a $3\times9$ grid are filled in using the numbers 1 through 9 so that each row contains 9 different numbers, and each of the three $3\times3$ blocks heavily outlined in the example below contains 9 different numbers, as in the first three rows of a Sudoku puzzle.

    | 4 | 2 | 8 | 9 | 6 | 3 | 1 | 7 | 5 | \\
    | 3 | 7 | 9 | 5 | 2 | 1 | 6 | 8 | 4 | \\
    | 5 | 6 | 1 | 8 | 4 | 7 | 9 | 2 | 3 | \\

    The number of different ways to fill such a grid can be written as $p^a\cdot q^b\cdot r^c\cdot s^d$, where $p,q,r,$ and $s$ are distinct prime numbers and $a,b,c,$ and $d$ are positive integers. Find $p\cdot a+q\cdot b+r\cdot c+s\cdot d$.\\[0.5em]

    \textbf{Baseline CoT.}
    The standard explicit CoT run explicitly enumerates the combinatorial factors. It first observes that the first row can be arbitrary ($9!$ choices). Given the first row, it derives $56$ possible set-level assignments for the second row and $6^3$ within-block orderings. It then notes that the third row still has independent within-block orderings, contributing another $6^3$ factor:
    $$
    N_{\mathrm{CoT}} = 9!\cdot 56\cdot 6^6 = 2^{16}3^{10}5^1 7^2.
    $$
    Therefore,
    $$
    2\cdot16 + 3\cdot10 + 5\cdot1 + 7\cdot2 = 81.
    $$
    The final prediction is correct:
    $$
    \mathrm{Pred}=81,\qquad \mathrm{Gold}=81.
    $$
    The important qualitative point is that the explicit trace keeps the row/block assignment and ordering factors separate until the final factorization.\\[0.5em]

    \textbf{Latent reasoning.}
    The output is organized as a latent segment followed by an explicit CoT segment. Abridged from the raw trace:
    $$
    ~\underbrace{\cdots}_{\substack{\text{low-entropy\ latent\ tokens\ }\\
    H_{\max}=1.625}}
    ~\texttt{<|endoftext|>}
    ~\texttt{<think>}
    $$
    Given Block A, the number of valid Block B's is $56\times216$ ... Block C is uniquely determined ... Therefore, the total number of grids is $9!\times56\times216$ ... Hence, the final answer is $66$.\\[0.5em]

    Thus the explicit CoT after the latent segment follows a superficially similar counting strategy, but it treats the third block as uniquely determined once the first two blocks are chosen. This makes the trace use
    $$
    N_{\mathrm{latent}} = 9!\cdot 56\cdot 6^3 = 2^{13}3^7 5^1 7^2
    $$
    instead of $9!\cdot56\cdot6^6$. This gives
    $$
    2\cdot13 + 3\cdot7 + 5\cdot1 + 7\cdot2 = 66.
    $$
    The final prediction is incorrect:
    $$
    \mathrm{Pred}=66,\qquad \mathrm{Gold}=81.
    $$
  \end{tcolorbox}

  \vspace{0.5em}

  \caption{Low-entropy latent reasoning can still fail. On AIME 2025 Problem~10, the baseline explicit CoT run answers correctly, whereas the LaTER run remains low-entropy during latent reasoning but converges to an incorrect answer. This suggests that entropy is not sufficient to certify correctness of compressed latent reasoning states.}
  \label{fig:failure_low_entropy_latent}
\end{figure}

\clearpage

\clearpage
\section{Licenses for Existing Assets}
\label{app:licenses}

The table below summarizes the external models, datasets, and software explicitly used in this paper. For each asset, we list the original owner, how it is used, and the released license or additional usage condition stated by the official model card, dataset card, or repository. We use these assets for research and evaluation, cite their original sources in the reference, and do not redistribute third-party assets outside their original release terms.

{\scriptsize
  \renewcommand{\arraystretch}{1.05}
  \setlength{\LTleft}{0pt}
  \setlength{\LTright}{0pt}
  \begin{longtable}{@{}>{\raggedright\arraybackslash}p{0.17\textwidth}>{\raggedright\arraybackslash}p{0.12\textwidth}>{\raggedright\arraybackslash}p{0.24\textwidth}>{\raggedright\arraybackslash}p{0.24\textwidth}>{\raggedright\arraybackslash}p{0.12\textwidth}@{}}
    \caption{Existing external assets used in this paper, with their original owners and released license or usage terms.}
    \label{tab:existing_assets_licenses}\\
    \toprule
    Asset & Type & Use in this paper & Original owner / credit & License \\
    \midrule
    \endfirsthead

    \toprule
    Asset & Type & Use in this paper & Original owner / credit & License \\
    \midrule
    \endhead

    \midrule
    \multicolumn{5}{r}{Continued on next page} \\
    \midrule
    \endfoot

    \bottomrule
    \endlastfoot

    Qwen3-14B & Model & Main backbone for training-free experiments and the initialized base model for trained LaTER & Qwen Team; \citet{yang2025qwen3technicalreport} & Apache-2.0. \\

    DeepSeek-R1-Distill-Llama-8B & Model & Training-free comparison model & DeepSeek-AI; \citet{guo2025deepseek} & MIT. \\

    OLMo-3-32B-Think & Model & Training-free comparison model & AI2 / OLMo Team; \citet{olmo2026olmo3} & Apache-2.0. \\

    AIME 2025 via MathArena & Benchmark & Main math evaluation benchmark & MathArena / ETH SRI Lab; \citet{balunovic_srimatharena_2025} & MIT. \\

    MATH-500 & Benchmark & Math evaluation benchmark & OpenAI / HuggingFaceH4; \citet{lightman2023lets} & MIT. \\

    Dolci-Think-SFT-32B & Dataset & Source dataset used to sample reasoning traces for constructing \textsc{Latent-Switch-69K} & AI2 / OLMo Team; \citet{olmo2026olmo3} & odc-by. \\

    GSM8K & Dataset & Math word-problem benchmark & OpenAI; \citet{cobbe2021trainingverifierssolvemath} & MIT. \\

    GPQA & Dataset & Science QA benchmark & \citet{rein2024gpqa} & MIT. \\

    ARC-Challenge & Dataset & Reasoning benchmark & AI2; \citet{allenai_arc} & MIT. \\

    HumanEval+ & Dataset & Code-generation benchmark & EvalPlus; \citet{evalplus} & Apache-2.0. \\

    MBPP+ & Dataset & Code-generation benchmark & EvalPlus; \citet{evalplus,evalperf} & Apache-2.0. \\

    FlashAttention-2 & Software & Attention kernel used in training implementation & Dao-AILab; \citet{dao2023flashattention2fasterattentionbetter} & BSD-3-Clause. \\

    \texttt{DeepSpeed} (ZeRO-3) & Software & Distributed training runtime & Microsoft / DeepSpeed Team; \citet{rajbhandari2020zero} & Apache-2.0. \\

  \end{longtable}
}


\end{document}